\documentclass[twoside,11pt]{article}

\usepackage{newtxtext,newtxmath}

\usepackage{graphicx}

\usepackage{setspace}

\usepackage{booktabs} 
\usepackage{tabularx} 
\usepackage{rotating} 
\usepackage[dvipsnames,table,xcdraw]{xcolor}

\usepackage[letterpaper,margin=0.8in]{geometry}

\renewenvironment{abstract}
	{\quotation}
	{\endquotation}

\date{}

\makeatletter
\renewcommand{\fnum@figure}{\textbf{Figure \thefigure}}
\renewcommand{\fnum@table}{\textbf{Table \thetable}}
\makeatother

\usepackage{scicite}

\usepackage{url}

\usepackage[colorlinks=true, linkcolor=blue, urlcolor=blue, citecolor=black]{hyperref}

\def\scititle{
    Embodied Intelligence for Sustainable Flight: A Soaring Robot with Active Morphological Control
}

\title{\bfseries \boldmath \scititle}

\author{
	Ghadeer Elmkaiel$^{1\ast}$,
	Syn Schmitt$^{2}$,
	Michael Muehlebach$^{1}$\\\and
	\small$^{1}$Learning and Dynamical Systems, Max Planck Institute for Intelligent Systems,\\
    \small 72076, Tübingen, Germany.\and
	\small$^{2}$Institute for Modelling and Simulation of Biomechanical Systems, University of Stuttgart,\\
    \small 70569, Stuttgart, Germany.\and
        \small $^\ast$Corresponding author. Email: ghadeer.elmkaiel@tuebingen.mpg.de\and
}

\begin{document} 

\maketitle

\begin{abstract} \bfseries \boldmath
Achieving both agile maneuverability and high energy efficiency in aerial robots, particularly in dynamic wind environments, remains challenging. Conventional thruster-powered systems offer agility but suffer from high energy consumption, while fixed-wing designs are efficient but lack hovering and maneuvering capabilities. We present Floaty, a shape-changing robot that overcomes these limitations by passively soaring, harnessing wind energy through intelligent morphological control inspired by birds. Floaty's design is optimized for passive stability, and its control policy is derived from an experimentally learned aerodynamic model, enabling precise attitude and position control without active propulsion. Wind tunnel experiments demonstrate Floaty's ability to hover, maneuver, and reject disturbances in vertical airflows up to $10~\mathrm{m/s}$. Crucially, Floaty achieves this with a specific power consumption of $10~\mathrm{W/kg}$, an order of magnitude lower than thruster-powered systems. This introduces a paradigm for energy-efficient aerial robotics, leveraging morphological intelligence and control to operate sustainably in challenging wind conditions.
\end{abstract}

\begin{figure}
	\centering
	\includegraphics[width=1.0\textwidth]{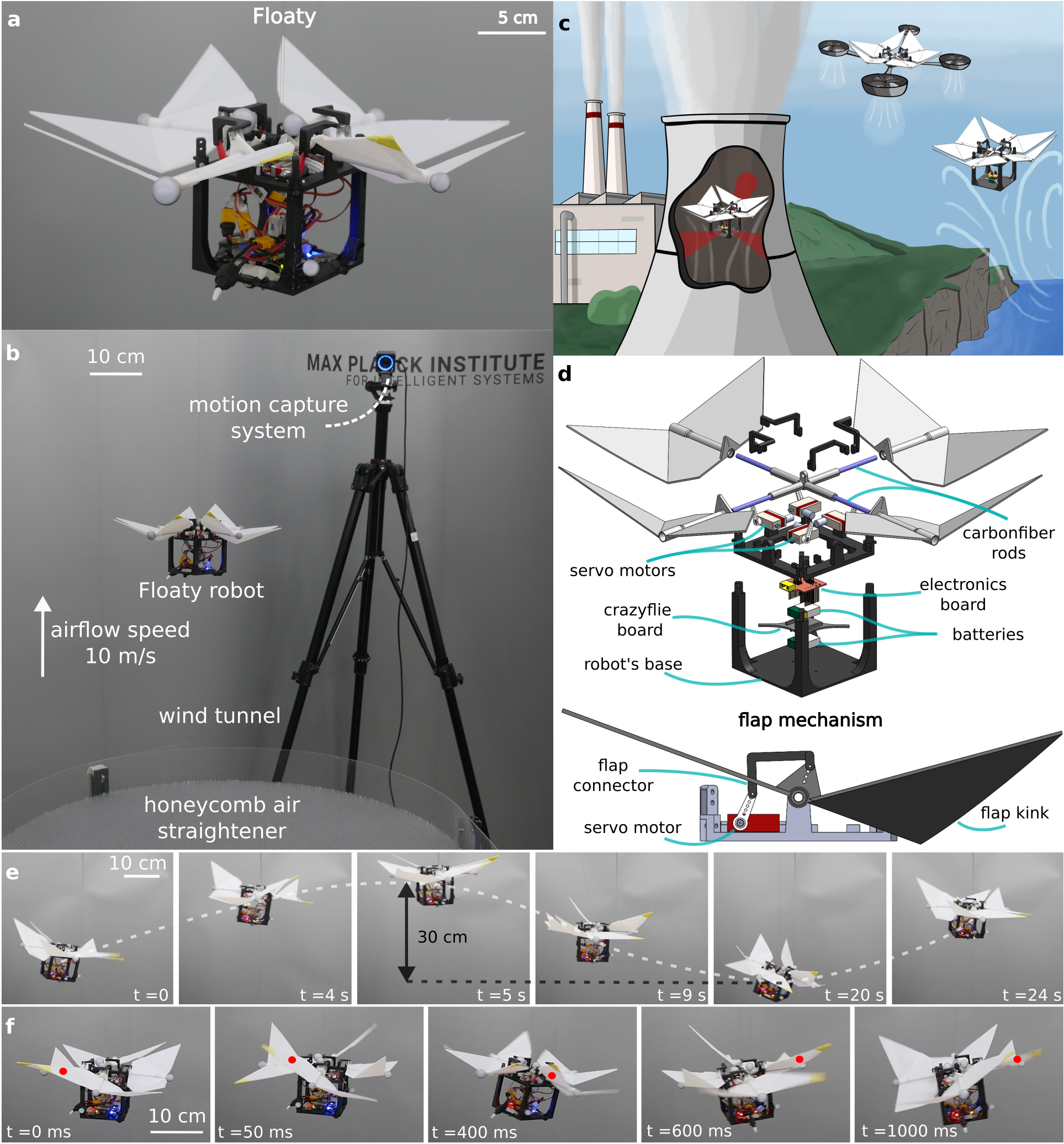} 

	\caption{\textbf{Floaty shape-changing soaring robot.}
		(\textbf{a}) Our robot “Floaty” flies in our work setup (\textbf{b}), consisting of the robot flying over the custom-built wind tunnel and the motion capture system cameras tracking the robot's flight. (\textbf{c}) Floaty's concept has the potential to be applied in factory smokestacks or in natural updrafts. (\textbf{d}) The robot can change its shape and configuration by rotating the four flaps, (\textbf{e} and \textbf{f}) controlling the robot's position and orientation.}
	\label{fig:main_figure} 
\end{figure}

\noindent
The development of autonomous aerial robots capable of sustained, agile, and energy-efficient operation in complex, dynamic environments is a central goal in aerial robot design \cite{ahmed2022recent, floreano2015science, wang2025symbiotic}. In nature, birds and insects exhibit remarkable mastery of flight in such conditions, inspiring researchers and engineers \cite{leutenegger2016flying, geronel2025overview, skydiving, warrick2009lift}. However, replicating this efficiency and adaptability, particularly in harnessing environmental energy sources like wind, remains a significant challenge for robotic systems \cite{bonnin2015energy}.\\
Aerial robotic paradigms face inherent trade-offs. Multirotor systems provide excellent hovering and maneuverability but are energetically inefficient due to their reliance on continuous thrust generation \cite{liu2017power, dietrich2017empirical, thibbotuwawa2019energy}. Fixed-wing aircraft offer energy-efficient flight but typically require forward velocity for lift and lack hovering capabilities \cite{boon2017comparison, panagiotou2020aerodynamic}. Bio-inspired flapping-wing robots \cite{ma2013controlled, jafferis2019untethered, gerdes2010review, delaurier1993aerodynamic, nguyen2018development, chin2020efficient, flapping_festo, flapping_fest_2, flapping_bird, Delft, Delft_2, Delft_3, jafferis2016non, karpelson2010energetics, chirarattananon2017dynamics, liang2023review} and morphing aircraft \cite{jeger2024adaptive, chang2024bird, ajanic2020bioinspired, di2017bioinspired} have shown promise in improving efficiency and flight performance, but typically lack agility during hover.
In parallel, research on passive flyers has explored the use of aerodynamic shaping to reduce energy consumption during descent or drift, as seen in wind-dispersed microfliers \cite{kim2021three, johnson2023solar, iyer2022wind} and samara-inspired autorotating devices \cite{win2021agile}. While these systems underscore the importance of aerodynamic design, they often lack fine active control or are restricted to unpowered descent or free-fall trajectories. Achieving sustained, controllable flight by harvesting energy solely from external airflow—without relying on onboard propulsion—remains an open challenge in aerial robotics (see Table~\ref{tab:vehicle_comparison}).\\
A key for closing this gap lies in morphological intelligence—designing robots whose physical shape leads to inherent stability—coupled with learned control strategies that enable precise and high maneuverability. Inspired by soaring birds that subtly adjust their wing and tail profiles to exploit updrafts \cite{swifts, kestrel, albatrosses, falcon, akos2010thermal, wilson1975sweeping}, we explore the design of a robot that actively modulates its aerodynamic shape to achieve hover and flight control using the energy from vertical wind, moving beyond passive descent.\\
We introduce Floaty, a passively propelled soaring robot designed to bridge the maneuverability-efficiency gap between fixed-wing aircraft and thruster-powered aerial vehicles (see Fig.~\ref{fig:main_figure}~a,~b and Supplementary \href{https://ghadeerelmkaiel.github.io/Floaty-Project/}{Video}~1). Unlike traditional multirotors that continuously expend energy to generate thrust, Floaty soars without active propulsion. Instead, it harnesses vertical wind flows (Fig.~\ref{fig:main_figure}~c) to enable energy-efficient hovering and agile flight. Floaty features four independently actuated flaps (Fig.~\ref{fig:main_figure}~d) that dynamically alter its aerodynamic profile. This morphological control allows it to modulate drag forces precisely, enabling control over its attitude and position (Fig.~\ref{fig:main_figure}~e~and~f, and Fig.~\ref{fig:control_concept}). The robot's design is optimized for passive stability, while its control policy is derived from an experimentally learned aerodynamic model, bridging the gap between the energy efficiency of passive gliders and the maneuverability of actively controlled systems.\\
Our main contributions are threefold:
(i) The design and implementation of a novel morphing robot capable of sustained, controlled passive soaring by dynamically controlling its shape to harness the wind energy.
(ii) A data-driven approach to model Floaty's complex aerodynamics from experimental data, enabling the derivation of an optimized and decoupled control strategy for agile maneuvering.
(iii) A significant reduction in flight energy consumption compared to conventional aerial systems, achieved through its intelligent morphological design (Fig.~\ref{fig:robustness}).
These advancements are realized through intelligent hardware and software co-design, optimizing Floaty's physical form and control system for wind-powered flight. The robot's capabilities, including efficient hovering and agile maneuvering, were comprehensively evaluated in our custom vertical wind tunnel (Fig.~\ref{fig:main_figure}~e~and~f), as detailed in this paper.

\section*{Results}

\subsection*{Flight principle using morphological control}

The robot achieves passive flight in vertically rising airflow environments, such as wind tunnels (Fig.~\ref{fig:main_figure}~b and Supplementary \href{https://ghadeerelmkaiel.github.io/Floaty-Project/}{Video}~1), without generating thrust.
This capability stems from an aerodynamically optimized body and a control policy that, in synergy, enable effective interaction with the surrounding air.
By dynamically rotating its four control surfaces—analogous to how birds adjust their wings and tail—the robot modulates its aerodynamic profile to manipulate drag forces.
While the robot’s ability to alter its shape is more limited than that of birds, this minimal morphological actuation is simple yet sufficient to enable stabilization, hovering, and maneuvering using only energy harnessed from the external wind, without relying on onboard propulsion.

\subsubsection*{Aerodynamic force generation}

The aerodynamic forces enabling Floaty's passive flight are primarily drag forces generated by its central base and four independently actuated flaps.
Therefore, the total force acting on the robot is modelled as the sum of the gravitational force ${F}_\mathrm{g}$ and the individual drag forces ${F}_\mathrm{D}$ exerted on each component of the robot's body (the base ${F}_\mathrm{B}$ and the four flaps ${F}_{i}$) as shown in equation (\ref{eq:total_force}):

\begin{equation}
    {F}_\mathrm{T} = {F}_\mathrm{g} + {F}_\mathrm{B} + \Sigma_{i=1}^{4}{F}_{i}.
    \label{eq:total_force}
\end{equation}

The base and the flaps are designed as thin plates; therefore, we use a simplified drag force model to describe the force ${F}_\mathrm{D}$ acting on each component (equation (\ref{eq:simplified_drag_force})). This model, detailed further in the Supplementary Information, is evaluated against experimental data and shows close alignment for operationally relevant flap angles (Fig.~\ref{fig:drag_model_evaluation}):

\begin{equation}
    F_{\mathrm{D}}(v_\mathrm{rel}, \theta)=\frac{\rho}{2} C_\mathrm{D,\perp} A \, v^2_{\mathrm{rel},{\perp}}(\theta),
    \label{eq:simplified_drag_force}
\end{equation}

where $\rho$ is the air density $\rho = 1.225\ \mathrm{kg/m^3}$, $A$ the total surface area of the flat plate, $C_\mathrm{D,\perp}$ the drag coefficient of the plate for flow perpendicular to the direction of relative motion, and $v_\mathrm{rel,\perp} (\theta)$ is the component of the relative velocity perpendicular to the plate surface, which is a function of the flap angle $\theta$. 

This model is the base for understanding how flap rotations translate into the forces required for flight control and is used to build a dynamical model of Floaty that closely aligns with experimental observations (see Supplementary Information for model details and evaluation).

\subsubsection*{Decoupled control through orthogonal actuation signals}
\begin{figure} 
	\centering
	\includegraphics[width=0.9\textwidth]{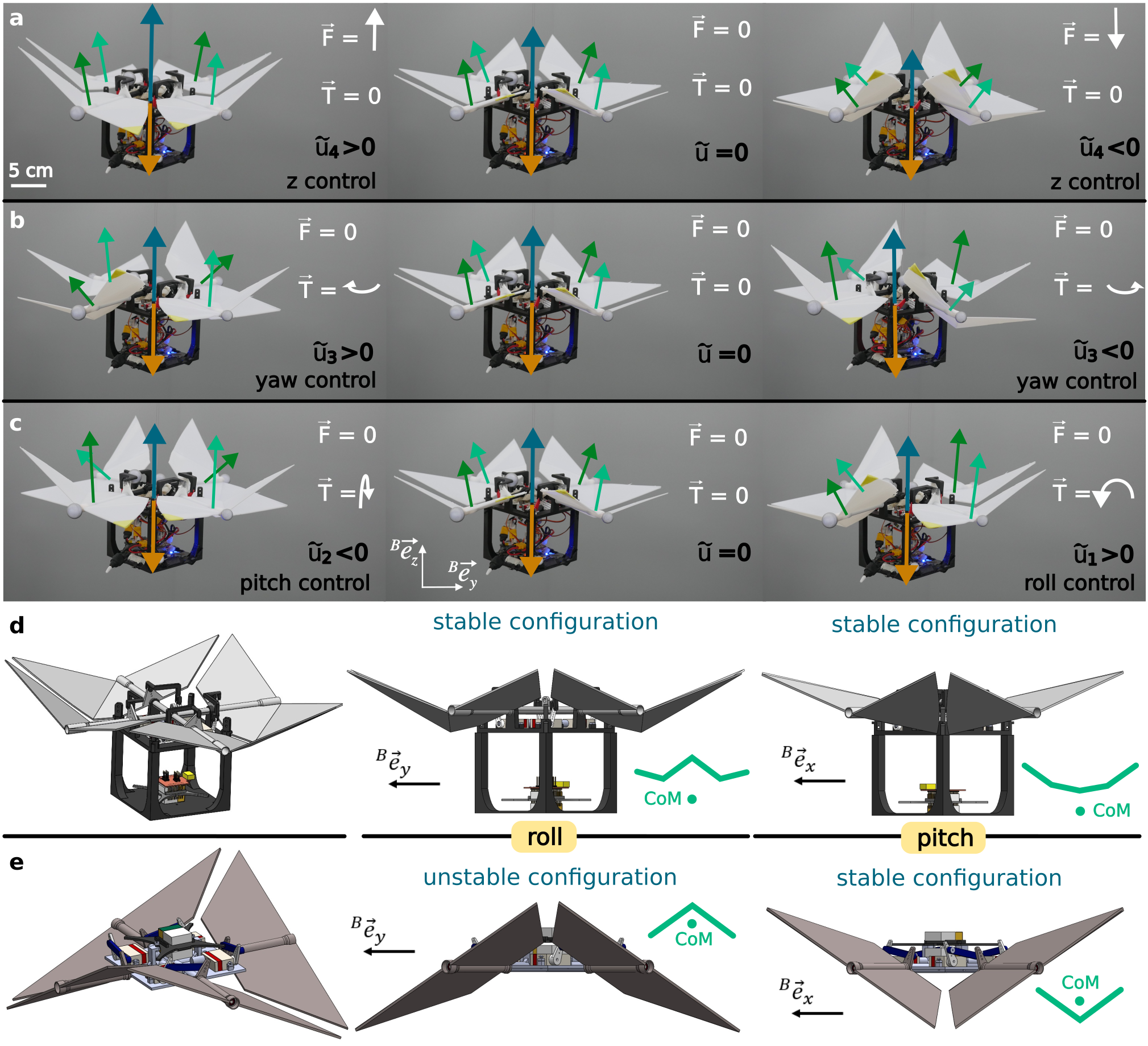} 
	\caption{\textbf{The concept of Floaty's control and design.}
		(\textbf{a}) The robot controls its $z$-acceleration by adjusting the size of the projected flap area. This is done by closing or opening all flaps simultaneously. In the neutral configuration (shown in the middle column), the combination of drag forces (shown in \textcolor{teal}{dark-blue}) counters gravity (shown in \textcolor{orange}{orange}).(\textbf{b}) Controlling yaw is achieved by opening two opposite flaps and closing the other two, creating a torque about the $z$-axis. (\textbf{c}) The robot controls roll and pitch by closing two adjacent flaps and opening the others, creating nonsymmetry and causing a torque about the $y$ or $x$-axis. (\textbf{d}) The kinks in the flaps and the lower center of mass “CoM”, enable stability of both roll and pitch dynamics. (\textbf{e}) Using flaps without kinks and higher CoM cause instability in the roll dynamics while the pitch dynamics might still be stable, depending on the height of the center of mass.}
	\label{fig:control_concept}
\end{figure}

To achieve precise and intuitive control over Floaty's six degrees of freedom using only these four actuated flaps, we designed a set of orthogonal compound control signals $(\tilde{u}_1, \tilde{u}_2, \tilde{u}_3, \tilde{u}_4)$ (Fig.~\ref{fig:control_concept} and Supplementary \href{https://ghadeerelmkaiel.github.io/Floaty-Project/}{Video}~2). These signals map desired changes in roll torque, pitch torque, yaw torque, and vertical force to specific, coordinated adjustments of the individual flap angles $(\theta_1, \theta_2, \theta_3, \theta_4)$ around a nominal hover configuration $\hat{\theta} = \hat{\theta}_\mathrm{H} = [\theta_\mathrm{H}, -\theta_\mathrm{H}, \theta_\mathrm{H}, -\theta_\mathrm{H}]^\top$ as defined in equation (\ref{eq:input_signals}) (see Supplementary Information for derivation details)
\begin{equation}
    \begin{bmatrix}
        \theta_1\\
        \theta_2\\
        \theta_3\\
        \theta_4
    \end{bmatrix}
    =
    \begin{bmatrix}
         \theta_\mathrm{H}\\
        -\theta_\mathrm{H}\\
         \theta_\mathrm{H}\\
        -\theta_\mathrm{H}
    \end{bmatrix}
    +
    \begin{bmatrix}
         1 &  1 & -1 & -1\\
         1 & -1 & -1 &  1\\
        -1 & -1 & -1 & -1\\
        -1 &  1 & -1 &  1
    \end{bmatrix}
    \begin{bmatrix}
        \tilde{u}_1\\
        \tilde{u}_2\\
        \tilde{u}_3\\
        \tilde{u}_4
    \end{bmatrix}.
    \label{eq:input_signals}
\end{equation}
This control allocation strategy, leveraging the robot's symmetry, aims to decouple the control inputs, enabling largely independent manipulation of key flight dynamics near hover.
The control signal $\Tilde{u}_4$ manages the total vertical force by altering the horizontal surface area of the flaps while maintaining a symmetrical configuration (Fig.~\ref{fig:control_concept}~a). The control signal $\Tilde{u}_3$ generates a torque about the $z$-axis by opening two opposite flaps while closing the other two (Fig.~\ref{fig:control_concept}~b), thereby controlling the $z$-torque. The signals $\Tilde{u}_1$ and $\Tilde{u}_2$ control the $x$ and $y$ torques, respectively, by closing two adjacent flaps while opening the other two (Fig.~\ref{fig:control_concept}~c). This asymmetry in flap angles increases drag on one side of the robot while decreasing it on the opposite side, thereby producing the desired torque on the robot.
Our flap-based control strategy is inspired by the way birds manipulate their wings and tail for agile flight control, enabling stabilization, hovering, and maneuvering using mainly the energy provided by the external wind.\\
While this actuation scheme provides control authority, achieving passively stable flight requires a careful optimization of the robot's design.

\subsection*{Embodied intelligence through design optimization}
Floaty does not generate thrust; however, it manipulates the magnitude and direction of aerodynamic forces through shape changes. Hence, the design of Floaty plays an essential role in defining its dynamics and flight characteristics. Our design strikes a balance between stability and controllability, while adhering to size constraints due to the wind tunnel environment, where our experiments are conducted.\\
The dynamics around hover, assuming vertical airflow, decouple into four parts, describing the roll, pitch, yaw, and vertical dynamics respectively.  This decoupling is facilitated by our control allocation, which enables independent manipulation of each part (see Supplementary Information and Fig.~\ref{fig:decoupled_dynamics}).
Specifically, $\Tilde{u}_4$ influences vertical dynamics ($\ddot{z}, \dot{z}, z$) as illustrated in Fig.~\ref{fig:control_concept}~a, and fig.~\ref{fig:decoupled_dynamics}~a, while $\Tilde{u}_3$ affects yaw dynamics ($\ddot{\gamma}, \dot{\gamma}, \gamma$) (Fig.~\ref{fig:control_concept}~b, and fig.~\ref{fig:decoupled_dynamics}~b). Similarly, $\Tilde{u}_1$ and $\Tilde{u}_2$ control roll ($\ddot{\alpha}, \dot{\alpha}, \alpha$) and pitch ($\ddot{\beta}, \dot{\beta}, \beta$) torques, respectively, also influencing lateral dynamics ($\ddot{y}, \dot{y}, y$) and ($\ddot{x}, \dot{x}, x$) (Fig.~\ref{fig:control_concept}~c, and fig.~\ref{fig:decoupled_dynamics}~c-d).
We denote the influence of one physical quantity on another using the symbol $f$ with double subscripts, where the first subscript indicates the affected quantity and the second indicates the influencing quantity (e.g., $f_{\ddot{\beta},\dot{x}}$ represents the effect of lateral velocity $\dot{x}$ on pitch acceleration $\ddot{\beta}$). Constants that directly link inputs to dynamics (e.g., $f_{\ddot{\beta},\Tilde{u}_2}$) are called \textit{input constants}, while others describing internal couplings (e.g., $f_{\ddot{\beta},\dot{x}}$) are \textit{structural constants}. 
The design of the robot determines the input and structural constants, which in turn affect the dynamics of the open-loop system.\\
A primary design challenge is to achieve passive open-loop stability, particularly for the roll dynamics. Initial investigations with flat flaps reveal that while the pitch dynamics exhibits inherent stability due to a “$\mathrm{V}$” shaped cross-section relative to the airflow (Fig.~\ref{fig:control_concept}~e, right), the roll dynamics presents an unstable "$\Lambda$" configuration (Fig.~\ref{fig:control_concept}~e, center). This results in a positive feedback structural constant linking roll angle to roll acceleration $f_{\ddot{\alpha},\alpha}>0$ (a positive value indicates instability) leading to an unstable pole (a root of the system's characteristic equation, $p_1\approx9.5$) in the open-loop system (initial model had $f_{\ddot{\alpha},\alpha}\approx+100~\mathrm{s}^{-2}$).\\
Inspired by how birds improve stability, two key modifications are implemented. First, the robot's center of mass (CoM) is lowered approximately $7~\mathrm{cm}$ below the plane of the flaps. Second, the flap geometry is altered by adding a $42.5^{\circ}$ kink on one side (Fig.~\ref{fig:control_concept}~d, Fig.~\ref{fig:design_and_electronics}~a). These values are optimized using our model to achieve inherent stability while maximizing control authority (Fig.~\ref{fig:robustness}~f~and~g). This transforms the roll axis cross-section from the unstable "$\Lambda$" shape (Fig.~\ref{fig:control_concept}~e, center) to the stable "$\_\Lambda\_$" shape (Fig.~\ref{fig:control_concept}~d, center). These two changes reverse the sign of the roll angle feedback, resulting in a negative structural constant ($f_{\ddot{\alpha},\alpha}\approx -16~\mathrm{s}^{-2}$), and shift the open-loop roll dynamics poles to stable locations (complex conjugate pair $p_{1,2}\approx-0.5\pm4\mathrm{i}$), as predicted by our dynamical model and empirically evaluated (see Methods and Table~\ref{tab:dynamic_consts} for estimation details of all structural and input constants). In contrast, the pitch dynamics remain stable, benefiting from the inherent "$\mathrm{V}$" configuration. Both roll and pitch exhibit negative angular rate feedback (structural constants $f_{\Ddot{\alpha},\Dot{\alpha}}<0$ and $f_{\Ddot{\beta},\Dot{\beta}}<0$), providing natural damping.\\
We optimize the nominal flap hover angle ($\theta_H$) to balance control authority across all axes. Smaller $\theta_H$ values improve yaw control but reduce pitch, roll, and height controllability, while very large angles diminish yaw authority. We find an optimal range of $20^\circ-25^\circ$, providing sufficient control authority (quantified by input constants 
$f_{\Ddot{\alpha},\tilde{u}_1}$, $f_{\Ddot{\beta},\tilde{u}_2}$, $f_{\Ddot{\gamma},\tilde{u}_3}$, and $f_{\Ddot{z},\tilde{u}_4}$ in Supplementary Table~\ref{tab:dynamic_consts}) for all three torques and the vertical force (Fig.~\ref{fig:robustness}~g).\\
Furthermore, a scale analysis is conducted to understand the influence of robot size and mass on maneuverability. Assuming a constant hover angle $\theta_H$, the analysis predicts that reducing the robot's scale enhances agility. Specifically, both (1) normalized linear acceleration-to-input ratio, ${\ddot{\tilde{x}}}/{\tilde{u}}$ (where $\ddot{\tilde{x}} =\ddot{x}/{l}$ is the linear acceleration $\Ddot{x}$ normalized by the robot's characteristic side length $l$) and (2) angular acceleration-to-input ratio, ${\dot{\omega}}/{\tilde{u}}$ (where $\dot{\omega}$ represents angular acceleration) increase with downscaling, implying that smaller robots can achieve equivalent relative maneuvers with less control input. Mass ($m$) scaling does not directly affect agility under these assumptions. However, it influences the required airflow speed shown in equation (\ref{eq:airflow_proportionality}): 
\begin{equation}
    v_\mathrm{air} \propto \sqrt{\frac{m}{l^2}},
    \label{eq:airflow_proportionality}
\end{equation}
and consequently the Reynolds number $\mathrm{R_e}$ (for our current design $\mathrm{R_e}\approx2\times10^5$) as expressed in equation (\ref{eq:Reynold_numbers}):
\begin{equation}
    \mathrm{R_e} = \frac{\rho v_\mathrm{air} l}{\mu} \propto\frac{\rho \sqrt{m}}{\mu},
    \label{eq:Reynold_numbers}
\end{equation}
where $\rho$ is the air density, and $\mu$ is the air’s kinematic viscosity.
This suggests that smaller, lighter designs are preferable for enhanced agility and operation in lower energy airflow environments, potentially favoring more laminar flow conditions. Detailed derivation of the scale analysis can be found in the Supplementary Information.

\begin{figure} 
	\centering
	\includegraphics[width=1.0\textwidth]{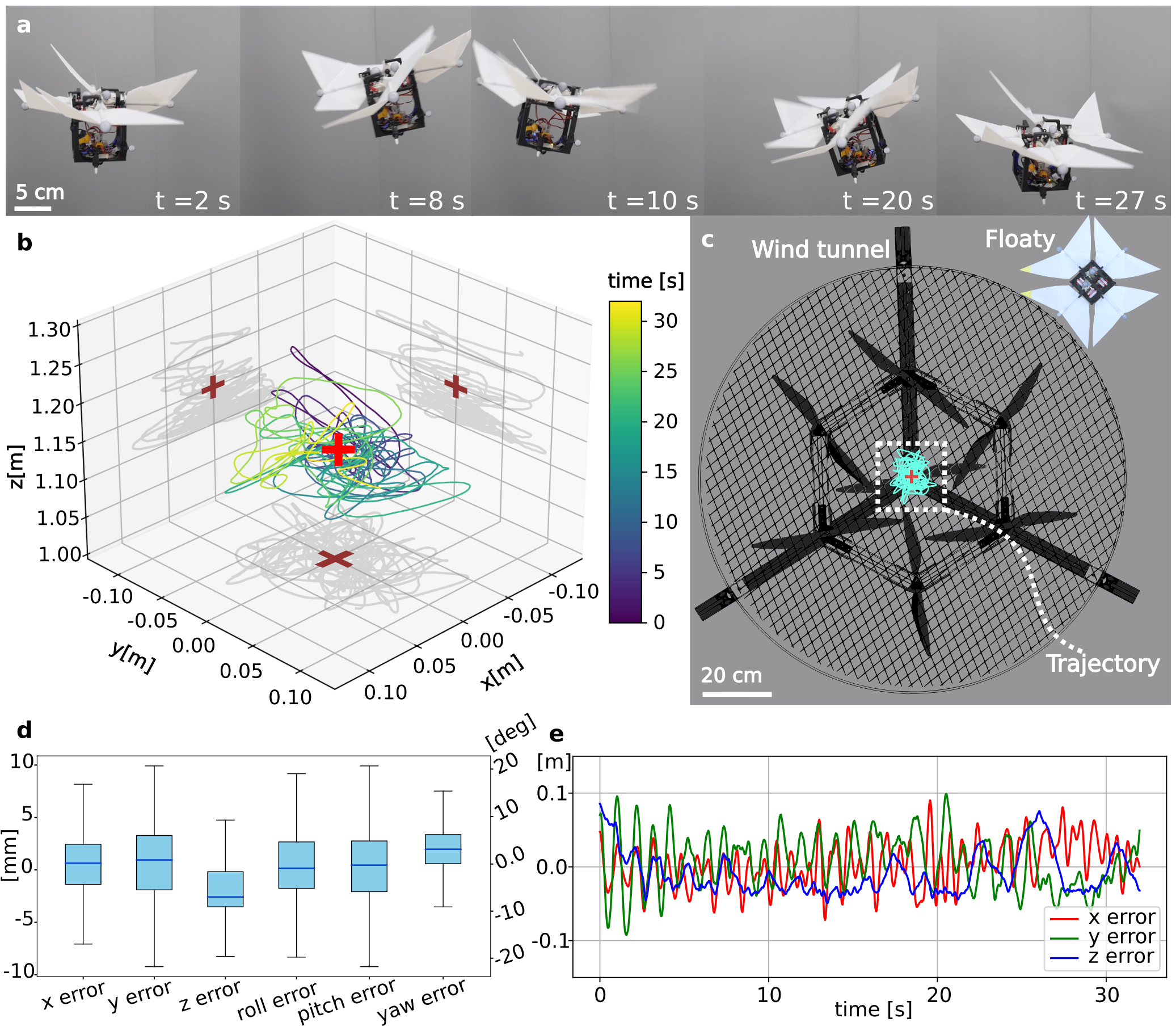} 

	\caption{\textbf{Hover experiment over the wind tunnel.}
		(\textbf{a}) The robot performs a hover experiment over the wind tunnel for $30~\mathrm{s}$. (\textbf{b} and \textbf{c}) The robot flies within a $20~\mathrm{cm}\times20~\mathrm{cm}\times20~\mathrm{cm}$ around the target position (shown in \textbf{\textcolor{red}{$+$}}), (\textbf{d} and \textbf{e}) with an error in the robot's position less than $6~\mathrm{cm}$ over each axis during most of the experiment, and an error in the attitude less than $20^\circ$.}
	\label{fig:Hover_figure} 
\end{figure}

\subsection*{Experimental evaluation of soaring capabilities}
To experimentally evaluate Floaty's capabilities as predicted by our analyses, we conducted a series of flight tests in our custom vertical wind tunnel (see Methods, Experimental Platform). 
Experiments were performed with a nominal vertical airflow speed of $v_\mathrm{air} = 10~\mathrm{m/s}$, corresponding to a Reynolds number ($\mathrm{R_e}$) of approximately $2\times10^5$ based on Floaty's characteristic length.
\subsubsection*{Hover performance}
We evaluate the robot's hovering capability by testing its ability to track a static target within the wind tunnel. The target position is set at $115~\mathrm{cm}$ above the center of the wind tunnel, and the experiment is conducted over a duration of 30 seconds, as illustrated in Fig.~\ref{fig:Hover_figure} and Supplementary \href{https://ghadeerelmkaiel.github.io/Floaty-Project/}{Video}~3. The robot hovers within a $20\times20\times20~\mathrm{cm}^3$ box centered around the target point, as shown in Fig.~\ref{fig:Hover_figure}~b-d. Observed positional deviations are primarily attributed to turbulence in the airflow. While oscillatory behavior is noted across $x$ and $y$-axes, the $y$-axis exhibits the largest deviations, indicating more pronounced oscillations in the $y$-dynamics compared to the other directions. 
This sensitivity can be attributed to specific aspects of the robot's y-axis dynamics (see Supplementary Information for details).
\subsubsection*{Yaw tracking performance}
We further evaluate the robot's yaw tracking performance while hovering. The experiment involves tracking wave signals with multiple frequencies and a fixed amplitude of ${\pi}/{2}~\mathrm{rad}$. The robot successfully tracks the target yaw angles with frequencies up to $0.15~\mathrm{Hz}$, achieving a maximum angular rate exceeding $80^\circ/\mathrm{s}$ while maintaining its position and stability. Each condition is tested over five trials, with results shown in Fig.~\ref{fig:experiments_fig}~b and Supplementary \href{https://ghadeerelmkaiel.github.io/Floaty-Project/}{Video}~4. Additionally, step response experiments are conducted (Fig.~\ref{fig:experiments_fig}~a and Supplementary \href{https://ghadeerelmkaiel.github.io/Floaty-Project/}{Video}~4), where the target yaw follows a step function with changes up to $110^\circ$. For step inputs, the robot achieves settling times of less than $0.5~\mathrm{s}$.
\subsubsection*{Vertical tracking performance}
We test the robot's ability to control its height during flight. In this experiment, the target height follows a sinusoidal signal with a frequency of $0.05~\mathrm{Hz}$ and an amplitude of $20~\mathrm{cm}$. The robot successfully tracks the target height while maintaining overall balance, as shown in Fig.~\ref{fig:experiments_fig}~c and Supplementary \href{https://ghadeerelmkaiel.github.io/Floaty-Project/}{Video}~5. The performance is repeatable and consistent across experiments, as indicated by showing results from five trials. It is noteworthy that the control signal for height ($z$-dynamics) has the lowest priority when calculating the flap angles from the input signals. The robot's attitude stabilization can override the height control signal, since height control is not critical for stability, which can be seen in the tracking data in Fig.~\ref{fig:experiments_fig}~c.
\subsubsection*{Horizontal tracking performance}
\begin{figure} 
	\centering
	\includegraphics[width=1.0\textwidth]{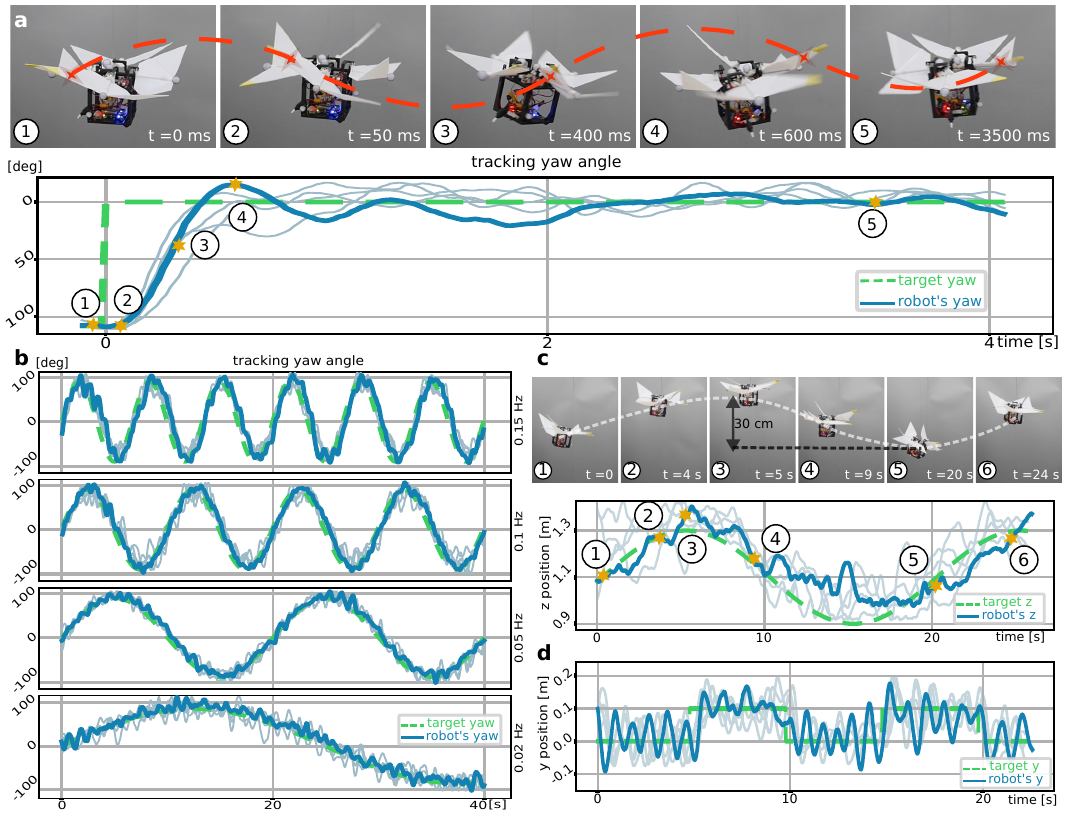}

	\caption{\textbf{Tracking robot position and yaw angle.}
		(\textbf{a})  The experiments include the analysis of yaw tracking with a step response with a yaw angle change over $100^\circ$, (\textbf{b}) and tracking a wave signal with an amplitude of $\pi/2$ and different frequencies [$0.02-0.15~\mathrm{Hz}$]. (\textbf{c}) The robot performs $z$-position tracking experiment with the target height following a sine wave with $20~\mathrm{cm}$ amplitude. (\textbf{d}) The robot performs $y$-position tracking with the target position following a square signal with $10~\mathrm{cm}$ amplitude. Each experiment was repeated multiple times, with individual trials shown in \textcolor[RGB]{157,186,198}{light-blue} and a representative trial highlighted in \textcolor[RGB]{19,129,178}{dark-blue}.}
	\label{fig:experiments_fig} 
\end{figure}

Similar to the height tracking experiment, we conduct multiple horizontal position tracking experiments. Although the hover experiments inherently include position tracking, this test highlights the robot's performance in following a dynamically changing $y$-position. Tracking performance in the $y$-direction is particularly informative, as this axis tends to exhibit more oscillatory behavior and is more sensitive to turbulence in the airflow.
The target $y$-position is set to follow a square wave signal with an amplitude of $10~\mathrm{cm}$ (Fig.~\ref{fig:experiments_fig}~d). The amplitude is constrained to $10~\mathrm{cm}$ due to the working area of our wind tunnel, where suitable airflow conditions for flight are confined to an effective cylindrical flight region of approximately $40~\mathrm{cm}$ in diameter (see Supplementary Information for wind tunnel characterization).

\subsubsection*{Energy efficiency via passive soaring}
\begin{figure} 
	\centering
	\includegraphics[width=1.0\textwidth]{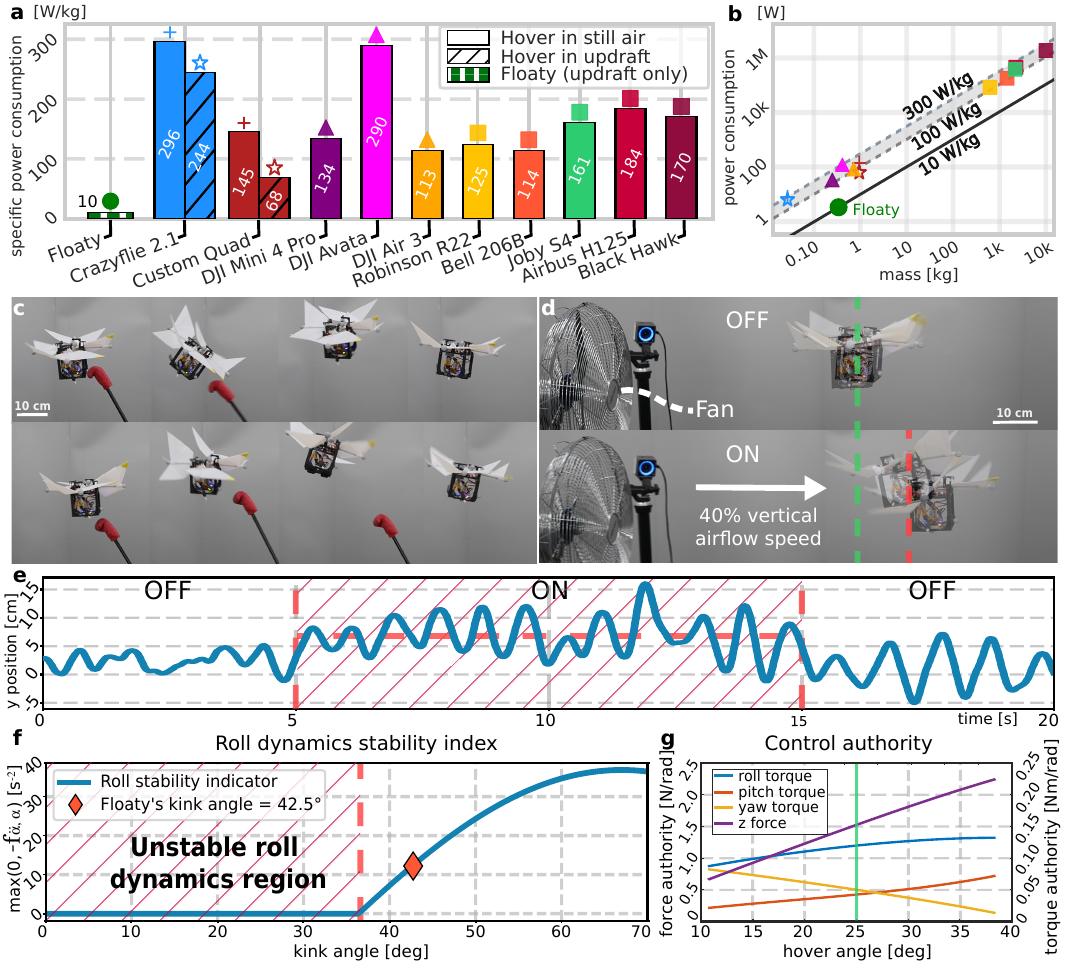}

	\caption{\textbf{Floaty's passive soaring enables remarkable energy efficiency and robust performance.}
        (\textbf{a}) Floaty's specific power consumption ($10~\mathrm{W/kg}$) is an order of magnitude lower than conventional aerial systems, as shown in a comparative analysis. Updraft conditions are indicated by a "Star" symbol.
    (\textbf{b}) Power consumption versus mass illustrates Floaty's significant deviation from the typical energy scaling (shaded in gray) of other aircraft.
    (\textbf{c}) Demonstration of robustness: Floaty recovering from direct physical pushes.
    (\textbf{d,~e}) Floaty maintains stability and control (\textbf{e}, y-position) when exposed to strong lateral wind (\textbf{d}, fan ON), equivalent to $40\%$ of the main vertical airflow. (\textbf{f}) The kink angle (\textcolor[RGB]{255,87,51}{$42.5^{\circ}$}) was optimized to achieve inherent roll stability with a margin for model error, and (\textbf{g}) the hover angle (\textcolor[RGB]{46,204,113}{$25^{\circ}$}) is optimized to maximize control authority over all degrees of freedom.}
	\label{fig:robustness} 
\end{figure}

A key benefit of Floaty's design concept is its energy efficiency, whereby the energy required for flight is harnessed from the vertical airflow. Unlike traditional motorized aerial vehicles where actuators (e.g., propellers and turbines) consume the majority of power, Floaty's servo actuators are only responsible for precisely rotating its flaps to modulate aerodynamic forces, consuming an order of magnitude less energy than propulsion systems (see Table~\ref{tab:vehicle_comparison}). The design places the center of pressure for each flap near its rotational axis, minimizing servo motor load and further contributing to this high energy efficiency.
To test this experimentally, we conduct continuous hovering endurance tests, in which the robot hovers over the wind tunnel until its two $250~\mathrm{mAh}$ single cell LiPo batteries (each with a capacity of $0.925~\mathrm{Wh}$) are depleted. Across repeated trials, flight durations consistently averaged approximately $33$ minutes. This corresponds to a total average power consumption of roughly $3.4~\mathrm{W}$, of which $0.3~\mathrm{W}$ is attributed to the onboard electronics (microcontroller running the flight controller and inertial sensors). Given Floaty's mass of $340~\mathrm{g}$, this translates to a specific power consumption of approximately $10~\mathrm{W/kg}$ during hover (Fig.~\ref{fig:robustness}~a~and~b).\\
This performance represents a significant improvement over conventional aerial systems. Multirotors, such as quadcopters or helicopters, typically exhibit specific power consumptions in the range of $100-250~\mathrm{W/kg}$ during hover \cite{dietrich2017empirical, dorling2016vehicle, liu2017power, helicopter, dji_mavic3pro_specs_2023, dji_air3_specs_2023, dji_avata_specs_2022, dji_mini4pro_specs_2023, prouty_helicopter_2005, faa_tcds_h2sw_bell206, faa_tcds_r22_robinson, easa_tcds_h125_airbus} (see Supplementary Information for detailed analysis of the energy efficiency of different flying vehicles). While operating in updrafts can reduce their power consumption \cite{leishman2006principles}, for updraft speeds comparable to our experimental conditions ($8$–$11~\mathrm{m/s}$), this improvement is often limited. For instance, even optimistic estimates suggest only up to a $60\%$ reduction in helicopter's power under such conditions (see Ch. 2.13 in Ref.~\cite{leishman2006principles}).\\ 
To provide a direct comparative context, we evaluated the power consumption during hover of two differently scaled quadcopters:\\
\textit{1. Micro quadcopter (Crazyflie):} A Crazyflie 2.1 quadcopter ($27~\mathrm{g}$), which utilizes the same core microcontroller as Floaty and is powered by a single $250~\mathrm{mAh}$ battery, consumes approximately $8~\mathrm{W}$ during hover in still air (specific power $\approx296~\mathrm{W/kg}$) \cite{giernacki2017crazyflie}. This is more than double Floaty's total power consumption, despite the Crazyflie being nearly 13 times lighter. When attempting to fly the Crazyflie in our wind tunnel's updraft, precise control proved challenging, even at reduced airflow speeds (see Fig.~\ref{fig:robustness}~a~and~b and Supplementary \href{https://ghadeerelmkaiel.github.io/Floaty-Project/}{Video}~6), and no significant improvement in its flight efficiency is observed.\\
\textit{2. Custom medium-scale quadcopter:} We also test a custom-built quadcopter with a larger frame size ($40\times40~\mathrm{cm}$) and a mass of $950~\mathrm{g}$, making its physical dimensions more comparable to Floaty's. In still air, this quadcopter exhibits a specific power consumption of $145~\mathrm{W/kg}$. When tested over the wind tunnel in an updraft of $\approx10~\mathrm{m/s}$, its specific power consumption is reduced to $68~\mathrm{W/kg}$, representing a $53\%$ improvement in energy efficiency.\\
Despite the efficiency gains observed for the custom quadcopter in an updraft, Floaty's specific power consumption of $10~\mathrm{W/kg}$ remains approximately seven times lower.\\
These comparisons, summarized in Fig.~\ref{fig:robustness}~a~and~b, starkly highlight the profound energy savings achieved by Floaty's passive soaring paradigm and intelligent morphological design.

\subsubsection*{Robustness and disturbance rejection} 
Floaty exhibits notable robustness to both physical external disturbances and variations in airflow conditions. This resilience is demonstrated experimentally through two types of tests:
First, the robot's ability to recover from direct physical perturbations is assessed. Floaty consistently regains and maintains stable hover after being subjected to manual pushes from various directions using a lightweight stick, provided Floaty remains within the wind tunnel's effective airflow region (Fig.~\ref{fig:robustness}~c and Supplementary \href{https://ghadeerelmkaiel.github.io/Floaty-Project/}{Video}~7). This indicates effective closed-loop stabilization against transient external forces.
Second, Floaty demonstrates resilience to variations in the airflow itself, maintaining stable flight in vertical updrafts ranging from $8$ to $11~\mathrm{m/s}$. This robustness extends to disruptions within the primary airflow, including partial temporary occlusions and, more significantly, the introduction of a controlled horizontal airflow component (Fig.~\ref{fig:robustness}~d~and~e Supplementary \href{https://ghadeerelmkaiel.github.io/Floaty-Project/}{Video}~7). The horizontal component is generated by a large side fan, creating crosswind velocities up to approximately $4~\mathrm{m/s}$ (equivalent to roughly $40\%$ of the main vertical airflow speed). Even under these challenging mixed-flow conditions, Floaty adapts its flap control and successfully maintains stable flight within the test volume, showcasing its ability to soar with moderate variations in both airflow direction and magnitude.

\section*{Discussion}
This work successfully demonstrates the effectiveness and significant potential of Floaty, a biologically inspired soaring robot engineered to achieve controlled, energy-efficient flight through intelligent morphological adaptation to vertical airflow environments. By utilizing its four independently actuated flaps to modulate aerodynamic drag forces—a principle directly inspired by the flight dynamics of birds—Floaty exhibits remarkable agility and an unprecedented level of energy efficiency for an actively controlled aerial system (see Table~\ref{tab:vehicle_comparison}). Comprehensive wind tunnel experiments have rigorously evaluated Floaty's capabilities, showcasing its ability to robustly maintain position and orientation during hover, accurately track dynamic positional and yaw trajectories, and effectively reject external disturbances. Crucially, these advanced flight characteristics are achieved with a specific power consumption of approximately $10~\mathrm{W/kg}$, an order of magnitude lower than conventional thruster-powered aerial systems. This efficiency stems directly from Floaty's passive harnessing of vertical airflow for lift generation, which drastically reduces the energy demands on its low-power servo actuators and highlights a paradigm shift away from energy-intensive propulsion.
\subsubsection*{Potential applications}
We present a design paradigm for flying robots capable of passively maneuvering and hovering in vertical airflow. An industrial example of such an environment is given by smokestacks in large factories (Fig. \ref{fig:main_figure}~c), where our design principle enables maneuvering inside the smokestack and performing regular scanning and observations while the factory is running. In factory smokestacks, the outflow velocities are typically in the range of $10-20~\mathrm{m/s}$ \cite{bryant2018improving, johnson2015design}, similar to the airflow conditions used in our experiments, suggesting that our robot could operate in such environments with minimal modifications.
Beyond industrial settings, naturally occurring vertical airflows—such as thermal updrafts, ridge updrafts, and atmospheric waves—offer another potential application for our design (Fig.~\ref{fig:main_figure}~c).
Additionally, our concept could be applied to rocket re-entry control and stabilization. By employing a similar flap mechanism, the design could facilitate precise attitude control during re-entry, leveraging aerodynamic forces to stabilize the vehicle. This concept is similar to the flaps used on SpaceX’s Starship, which aid in stabilizing the rocket during descent before executing the belly-flop maneuver \cite{starship2022}. Similarly, the flap mechanism could be integrated into weather balloon sensor systems, guiding the sensor payload during descent, making recovery process easier. Furthermore, our approach could enhance the energy efficiency of existing unmanned aerial vehicles by optimizing their ability to harness vertical airflow (see Fig.~\ref{fig:main_figure}~c). This highlights Floaty’s adaptability to a diverse range of applications.
\subsubsection*{Limitations and future work}
Despite the advantages of Floaty's concept, several limitations remain. In soaring scenarios with vertical airflow, the flap hover angle can be adjusted to accommodate wind speed variations. However, this adaptability has limits.
For significantly different airflow speeds—particularly above $11~\mathrm{m/s}$ or below $8~\mathrm{m/s}$—modifications to the robot’s size and/or weight are necessary to maintain hover.
While Floaty demonstrates robustness against external disturbances and lateral winds in our experiments, the design cannot handle situations where the horizontal component of the airflow becomes dominant. In such conditions, the current design causes the robot to drift, as it lacks the means to generate sufficient lateral control forces.\\
This work represents a step toward developing energy-efficient aerial vehicles. Future research will explore integrating Floaty's passive soaring capabilities with active propulsion systems. Such hybrid designs could combine high efficiency with enhanced control authority, enabling operation across a broader range of wind conditions and application scenarios.
%
\section*{Methods}
\subsection*{Experimental platform}
Our experimental platform comprises four main subsystems: the Floaty robot, a custom vertical wind tunnel, an OptiTrack motion capture system, and an off-board control station.
The wind tunnel is a custom-built cylindrical setup with a diameter of $1.2~\mathrm{m}$. It uses seven U5 KV400 brushless motors equipped with $406~\mathrm{mm}$-diameter T-motor carbon fiber propellers to generate a vertical airflow with average velocities ranging from $8~\mathrm{m/s}$ to $11~\mathrm{m/s}$. In our experiments, we use an airflow velocity of $10~\mathrm{m/s}$.\\
Robot tracking during flight is achieved by an OptiTrack motion capture system \cite{optitrack2024} equipped with five cameras. The system captures the robot's position and attitude at a frequency of $200~\mathrm{Hz}$, relaying this information to the off-board system via ethernet.\\
The off-board system integrates pose data from the OptiTrack system and the robot, and calculates the robot's trajectory and its deviations from the desired path. This information is sent to the robot via radio communication with a frequency of $50~\mathrm{Hz}$. Simultaneously, the off-board system records incoming data from the robot for analysis.\\
The robot integrates the information it receives from the off-board system with the sensory information it gathers on-board from the inertial measurement unit and gyro sensors. The robot uses this information to generate corrective flap commands with a frequency of $100~\mathrm{Hz}$. A detailed description for each of the subsystems is in the Supplementary Information.
\subsection*{Robot design and fabrication}
Floaty's frame is designed to meet a combination of low mass and structural robustness, primarily using 3D printing for its fabrication.
The body frame is printed using fiber-reinforced filament (Onyx, Markforged) with a base thickness of $3~\mathrm{mm}$. The four flaps were printed from polylactic acid tough filament (PLA Tough, Bambu Lab). Each flap is $1.6~\mathrm{mm}$ thick and weighs approximately $25~\mathrm{g}$;  collectively, the flaps constitute about one-third of the robot's total mass of $340~\mathrm{g}$ (total flap mass$\approx100~\mathrm{g}$). Lightweight carbon fiber rods ($8~\mathrm{mm}$ diameter, $96~\mathrm{mm}$ length) serve as rotation axes for the flaps.
Flap actuation is achieved using four miniature $9~\mathrm{g}$ servo motors (KST X08H Plus), each connected to a flap via a parallelogram linkage mechanism (fig.~\ref{fig:design_and_electronics}). Power for the entire system is supplied by two lightweight ($7.2~\mathrm{g}$ each), single cell $250~\mathrm{mAh}$ LiPo batteries connected in series to provide $7.6~\mathrm{V}$. 
The onboard controller is an adapted Crazyflie 2.1 board \cite{giernacki2017crazyflie}, with a custom electrical board for voltage regulation ($3.3~\mathrm{Volt}$ for Crazyflie) and servo signal distribution. 
Further details about the robot's design are provided in the Supplementary Information.
\subsection*{System identification and control}
In our physical model, we simplify the calculation of the drag force on the flaps by projecting the airflow to be perpendicular to the flap's surface. This assumption simplifies the drag force computation for various flap angles, as both the effective area and the drag coefficient $C_\mathrm{D}$ in the drag force equation become constants, independent of the flap angle as shown in equation (\ref{eq:flaps_drag_force_simplified}). This model is evaluated through multiple experiments, described in detail in the Supplementary Information. The model serves as a foundation for optimizing the shape and design of the robot and designing a preliminary controller.
An initial stabilizing controller allowed for brief hovering flights, during which experimental data is collected to identify the robot's closed-loop dynamics.
Using the measured parameters of the closed-loop dynamics (parameters in supplementary Table \ref{tab:dynamic_consts}), we estimated the open-loop system dynamics and iteratively refined the control policy to enhance flight performance.
We use a proportional controller with an integrator term only for the $z$-position control. Our robot uses the FreeRTOS operating system, which allows for non-sequential, asynchronous processes. The control process runs with a frequency of $100~\mathrm{Hz}$ and it updates the flaps commands using the latest state estimate. The state vector used in our control consists of 16 values, three for position, three for linear velocity, three for orientation, three for angular velocity, and four for the estimated flaps angles. State estimation is performed via a Kalman filter. The angular velocity is updated using the onboard gyro sensor with a frequency of $100~\mathrm{Hz}$, which also updates the orientation estimate. The position and orientation values are updated using the OptiTrack tracking data, which is sent to the robot via radio using the Crazyflie radio dongle, with a frequency of $50~\mathrm{Hz}$. We note that the information flow from the OptiTrack to the off-board system to the robot is not instant, and has a delay of about $40~\mathrm{ms}$, which we compensate for with a forward prediction method using the linearized dynamical model of our robot.
\subsubsection*{Data-driven estimation of dynamics constants}
To estimate the open-loop dynamics constants, we first identified the closed-loop system characteristics under two distinct preliminary linear controllers.
As a result, the system characteristics, such as the angle feedback constants (shown in Fig.~\ref{fig:decoupled_dynamics}) can be estimated from data.
While it is theoretically possible to estimate these constants directly by observing the robot's behavior in a wind tunnel without any controller, practical challenges arise. For instance, the robot would need to be flying untethered to ensure the data reflects its natural dynamics accurately, which is challenging to achieve.
Therefore, to measure the system's feedback constants, we used a preliminary linear controller that can achieve basic short flight and allowed us to record mid-flight data. Using the data, we estimate the closed and open-loop dynamics. In the case of the roll dynamics, the system's roll acceleration $\Ddot{\alpha}$ can be described by equation (\ref{eq:roll_dynamics_general}):
\begin{equation}
    \Ddot{\alpha} = a \Tilde{u}_{1,1} + b\dot{\alpha} + c \alpha + d \dot{y} + N,
    \label{eq:roll_dynamics_general}
\end{equation}
where $a=f_{\Ddot{\alpha},\Tilde{u}_{1,1}}, ~b = f_{\Ddot{\alpha},\Dot{\alpha}}, ~c = f_{\Ddot{\alpha},\alpha}, ~d = f_{\Ddot{\alpha},\dot{y}}, \Tilde{u}_{1,1}$ is the specific control input for this controller variant, and $N$ is the noise signal. 
To control the roll dynamics, we designed a linear proportional controller for the control input $\Tilde{u}_{1,1}$. This controller is first optimized in a physics simulation and subsequently fine-tuned on the real robot to meet performance requirements. The controller takes the following form:

\begin{equation}
    \Tilde{u}_{1,1} = \Bar{b}_1\dot{\alpha} + \Bar{c}_1 \alpha + \Bar{d}_1 \dot{y},
    \label{eq:roll_dynamics_control_signal}
\end{equation}
where $\Bar{b}_1, \Bar{c}_1$, and $\Bar{d}_1$ are the tuned parameters corresponding to damping, stiffness, and lateral velocity coupling, respectively. 
Therefore, we can write the roll acceleration as follows:
\begin{equation}
    \Ddot{\alpha} = (a\Bar{b}_1 +b)\dot{\alpha} + (a\Bar{c}_1 +c)\alpha + (a\Bar{d}_1 +d)\dot{y} + N.
    \label{eq:roll_dynamics_expanded}
\end{equation}
We record the values $\dot{\alpha}, \alpha,$ and $\dot{y}$ during flight with the motion capture system, and transform the relevant time-series data $\alpha(t), \dot{\alpha}(t),$ and $\dot{y}(t)$ to the frequency domain ${A}(\mathrm{j}\omega), \mathrm{j}\omega A(\mathrm{j}\omega),$ and $\mathrm{j}\omega {Y}(\mathrm{j}\omega)$. This transformation allows the application of a low pass filter, which removes most of the noise signal $N$. In the frequency domain, we can estimate the roll acceleration using the roll rates

\begin{equation}
    \Ddot{A}(\mathrm{j}\omega) = \mathrm{j}\omega\dot{A}(\mathrm{j}\omega).
    \label{eq:roll_dynamics_frequency}
\end{equation}
We then solve the system of equations (using the low frequencies):

\begin{equation}
    \mathrm{j}\omega\dot{A}(\mathrm{j}\omega)=[\dot{A}(\mathrm{j}\omega) ~A(\mathrm{j}\omega) ~\dot{Y}(\mathrm{j}\omega)][b^\ast_1 ~c^\ast_1 ~d^\ast_1]^\top,
    \label{eq:roll_dynamics_sys_of_equations}
\end{equation}
and find the closed-loop dynamics constants $b^\ast_1=(a\Bar{b}_1 +b), ~c^\ast_1=(a\Bar{c}_1 +c), ~d^\ast_1=(a\Bar{d}_1 +d)$.\\
We repeat the process using a different controller $\Tilde{u}_{1,2} = \Bar{b}_2\dot{\alpha} + \Bar{c}_2 \alpha + \Bar{d}_2  \dot{y}$ and we find the new closed-loop dynamics constants $b^{\ast}_2=(a\Bar{b}_2 +b), ~c^\ast_2=(a\Bar{c}_2 +c), ~d^\ast_2(a\Bar{d}_2 +d)$. We calculate the input constant $a$ as follows:

\begin{equation}
    a=\frac{b^\ast_1 - b^\ast_2}{\Bar{b}_1 - \Bar{b}_2} = \frac{c^\ast_1 - c^\ast_2}{\Bar{c}_1 - \Bar{c}_2}.
    \label{eq:roll_dynamics_input_const}
\end{equation}
Using the estimated input authority $a$ and the identified closed-loop dynamics constants $b^{\ast}, ~c^{\ast},$ and $d^{\ast}$ we can calculate the open-loop structural constants $b, ~c,$ and $d$.\\
In our experiments, we employed two different controllers with $\Bar{c}_1 = -0.2$ and $\Bar{c}_2 = -0.15$, and measured the corresponding closed-loop dynamics constants as $c^\ast_1=-60$, $c^\ast_2=-48.5$. From these values, we estimated $a = (c^\ast_1 - c^\ast_2)/(\Bar{c}_1 - \Bar{c}_2) = 230$.\\
Subsequently, we calculated $c=c^\ast_1 -a\Bar{c}_1=-16$. The negative value of $c=f_{\ddot{\alpha}, \alpha}=-16$ confirms the inherent structural stability of the roll dynamics. This is what we expect according to the design optimization ($f_{\ddot{\alpha}, \alpha}\approx-14$ in Fig.~\ref{fig:robustness}~g).\\
Additionally, the input authority over the roll dynamics, represented by $a=f_{\ddot{\alpha}, \Tilde{u}_1}=230$, is an order of magnitude larger than the angular feedback effect, represented by the structural constant $c$ . This disparity ensures controllability of the roll dynamics.\\
For the pitch dynamics, we derived the corresponding values shown in Table \ref{tab:dynamic_consts}. The structural constant $c=f_{\ddot{\beta}, \beta}=-1560$ ensures stability in the pitch dynamics. Furthermore, while the input effect on the pitch dynamics $a=f_{\ddot{\beta}, \Tilde{u}_2}=-1066$ is slightly smaller than the angular effect, it is of a similar magnitude. This indicates that the pitch dynamics remain controllable despite the physical constraints on the maximum and minimum flap angles. Using a similar method, we estimate all other dynamics constants as described in Table \ref{tab:dynamic_consts}.

\newpage

\begin{sidewaystable}[htbp]
\centering
\caption{Comparison of aerial vehicle design paradigms}
\label{tab:vehicle_comparison}

\begingroup 

\setstretch{0.9}

\footnotesize 
\begin{tabularx}{\textwidth}{@{}lXXXXX@{}}

\toprule
\textbf{Characteristic} & 
\textbf{Thruster-powered} & 
\textbf{Fixed-wing} \newline (e.g., gliders) & 
\textbf{Flapping-wing robots} & 
\textbf{Passive drifters/fliers} & 
\textbf{Floaty (this work)} \\
\midrule

Propulsion method & 
Active thrust (propellers) & 
Aerodynamic lift & 
Active wing flapping & 
Passive (drag / autorotation) & 
\textbf{Passive soaring} \\ 
\addlinespace

Hovering capability & 
Yes (excellent) & 
No & 
Yes (in many designs) & 
No (uncontrolled descent) & 
\textbf{Yes (in vertical airflow)} \\
\addlinespace

Maneuverability & 
High & 
Low &  
High &  
Very low / none & 
\textbf{High / agile} \\
\addlinespace

Energy efficiency & 
Low (100--250 W/kg) & 
High (gliding) & 
Low (up to 500 W/kg) & 
N/A (passive) & 
\textbf{High ($\approx$10 W/kg)} \\
\addlinespace

Wind harnessing & 
Limited & 
Yes (for soaring/gliding) & 
Limited & 
Yes & 
\textbf{Yes} \\ 
\addlinespace

Control authority & 
High & 
Good, but no hover control & 
High, but often complex & 
Very low / none & 
\textbf{High (for attitude \& position)} \\
\addlinespace

Key advantage & 
High agility and hovering in any condition & 
High energy efficiency for long-range flight & 
Potential for high efficiency and agility & 
No energy needed for propulsion & 
\textbf{High efficiency \& maneuverability} \\
\addlinespace

Key limitation & 
High energy consumption & 
Inability to hover or fly at low speeds & 
Mechanically complex; high energy consumption & 
Lacks control; restricted to descent & 
\textbf{Requires external vertical airflow} \\
\addlinespace

Examples & 
\cite{dji_air3_specs_2023, dji_avata_specs_2022, dji_mavic3pro_specs_2023, dji_mini4pro_specs_2023, giernacki2017crazyflie, prouty_helicopter_2005, faa_tcds_h2sw_bell206} & 
\cite{boon2017comparison, panagiotou2020aerodynamic} & 
\cite{ma2013controlled, jafferis2019untethered, gerdes2010review, delaurier1993aerodynamic, nguyen2018development, chin2020efficient, flapping_festo, flapping_fest_2, flapping_bird, Delft, Delft_2, Delft_3, jafferis2016non, karpelson2010energetics, chirarattananon2017dynamics, liang2023review} & 
\cite{kim2021three, johnson2023solar, iyer2022wind, win2021agile} & 
\textbf{Floaty} \\

\bottomrule
\end{tabularx}

\endgroup 

\end{sidewaystable}


\clearpage 
\bibliography{bibliography}
\bibliographystyle{sciencemag_edited}


\section*{Acknowledgments}
The authors thank the International Max Planck Research School for Intelligent Systems (IMPRS-IS) for supporting Ghadeer Elmkaiel. We also thank Felix Grimminger for his advice and help during the initial phase of the project development, and we thank Felix Gr\"uninger for his advice and help with the materials used in the robot.
\paragraph*{Funding:}
This work was supported by the German Research Foundation and the Branco Weiss Fellowship, administered by ETH Zurich.

\newpage


\renewcommand{\thefigure}{S\arabic{figure}}
\renewcommand{\thetable}{S\arabic{table}}
\renewcommand{\theequation}{S\arabic{equation}}
\renewcommand{\thepage}{S\arabic{page}}
\setcounter{figure}{0}
\setcounter{table}{0}
\setcounter{equation}{0}
\setcounter{page}{1} 


\begin{center}
\section*{Supplementary Information}
\end{center}

\subsection*{Simplified drag model}
The drag force $F_{\mathrm{D}}$ measured on a body in a air with a relative motion between the body and the air can be written as equation (\ref{eq:drag_force_no_simplification}):
\begin{equation}
F_{\mathrm{D}}=\frac{1}{2}\rho C_\mathrm{D} A_{\perp} v_{\mathrm{rel}}^2,
\label{eq:drag_force_no_simplification}
\end{equation}
where $\rho$ is the air density $\rho = 1.225\ \mathrm{kg/m^3}$, $A_{\perp}$ the surface area of the object perpendicular to the relative movement direction, $C_\mathrm{D}$ the drag coefficient of the body, which depends on the body's orientation, and $v_\mathrm{rel}$ the relative speed between the object and the air.
The drag force depends solely on the relative velocity, provided that the object's orientation remains constant with respect to the airflow—meaning the portion of the object facing the airflow does not change. However, when the body's orientation changes, the equation becomes more complex. Both the drag coefficient, $C_\mathrm{D}$, and the area facing the airflow, $A_{\perp}$, become functions of the object's orientation, $\theta$:
\begin{equation}
F_{\mathrm{D}}(v_\mathrm{rel}, \theta)=\frac{\rho}{2} C_\mathrm{D}(\theta) A_{\perp}(\theta) v_{\mathrm{rel}}^2.
\label{eq:drag_force_with_orientation}
\end{equation}
While calculating $A_{\perp}(\theta)$ is relatively straightforward, determining $C_\mathrm{D}(\theta)$ is more challenging. Since our experiment requires an understanding of drag for various orientations, we developed a simplified model to estimate the drag force on thin flat plates, accounting for the plate's orientation.
For a thin flat plate in air, we make the following assumption:\\
\textbf{Assumption.} The drag force acting on a thin, flat plate is only due to the perpendicular component of the relative motion between the plate and the air.\\
This assumption allows us to project the relative airflow onto the normal of the plate's surface and calculate the drag force as if the plate were oriented directly into the airflow, with a relative speed of $v_{\mathrm{rel},\perp}$.
The advantage of this model is that by neglecting the effect of the parallel airflow component, we treat both the area facing the airflow and the drag coefficient as constants, independent of the plate's orientation. The only orientation-dependent variable is the projected perpendicular relative velocity, which is calculated using the plate’s orientation, speed, and the airflow vector. This simplification is adopted for initial model development, focusing on the dominant drag effects for thin, plate-like structures at the relevant flap angles encountered during controlled soaring.

\begin{figure}
	\centering
	\includegraphics[width=0.8\textwidth]{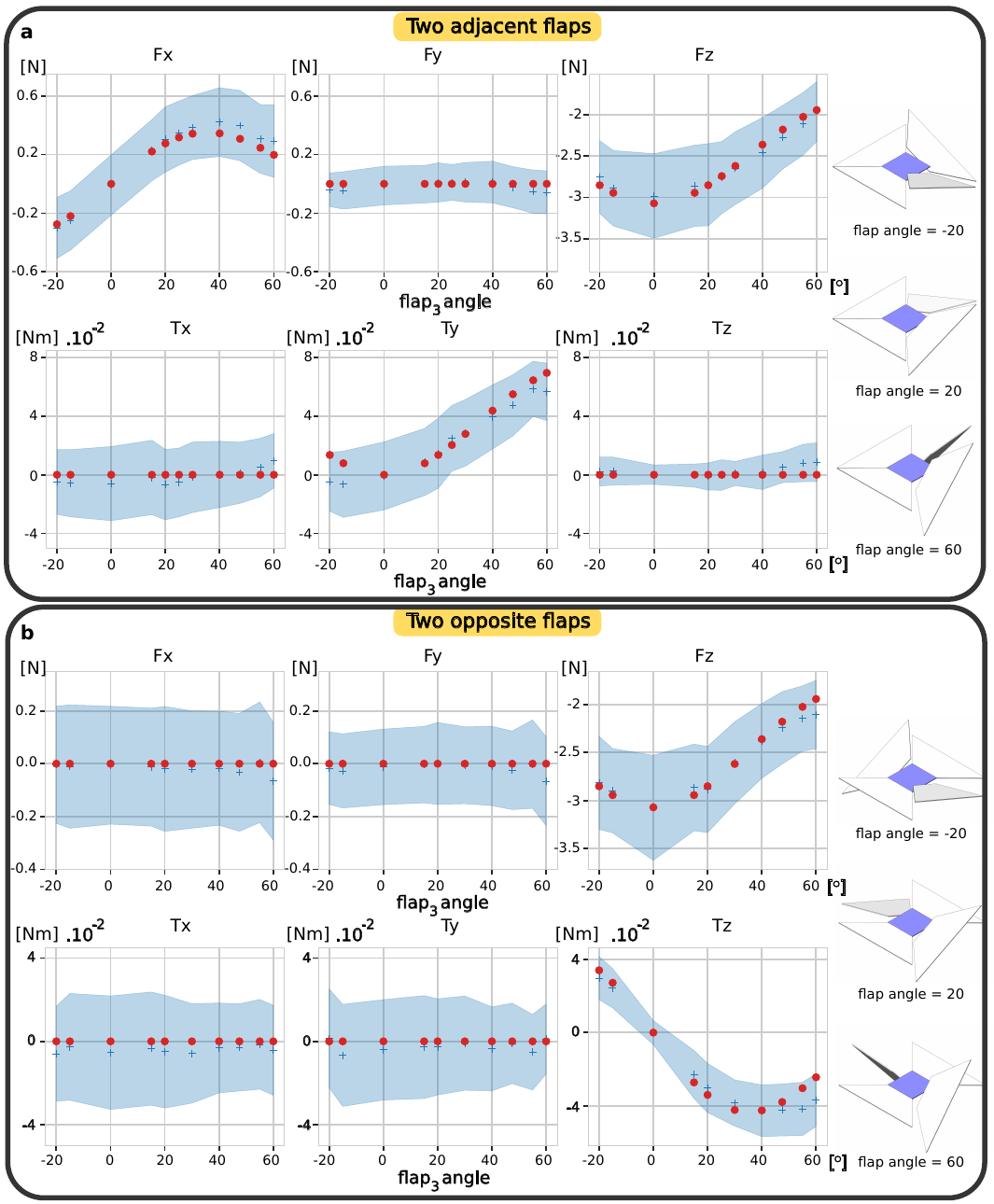} 

	\caption{\textbf{Simplified drag model evaluation.}
		(\textbf{a}) The experimental results for evaluating the simplified drag model in the case of moving flaps 2 and 3 with the same angle in different directions. The first row shows the forces while the second row shows the torques over $x$, $y$, and $z$-axes respectively. The (\textcolor{red}{\textbf{.}}) symbols show the model's predictions, while the (\textcolor{blue}{\textbf{+}}) symbols show the average measurements. The shaded area represents the standard deviation over the measurements. (\textbf{b}) The experimental results for evaluating the simplified drag model in the case of moving two opposite flaps (1 and 3) with the same angle in the same directions.}
	\label{fig:drag_model_evaluation} 
\end{figure}

\subsection*{Drag model evaluation}
We evaluate our drag model, and we check its accuracy experimentally. For this experiment, we use a robot with flat flaps. The robot is mounted on an ATI mini40 force and torque sensor \cite{ati_mini40} over the wind tunnel. Multiple measurements are taken for different static configurations. We test the validity of our model for multiple combined flap movements. For each configuration, $35000$ force and torque measurements are sampled over $10~\mathrm{s}$. The model is evaluated over multiple experiments such as, moving two adjacent flaps simultaneously (Fig.~\ref{fig:drag_model_evaluation}~a), and moving two opposite flaps simultaneously (Fig.~\ref{fig:drag_model_evaluation}~b). Results show that the simplified drag model provides accurate predictions across a broad range of flap angles near the nominal hover configuration. Notably, discrepancies arise only at extreme flap angles, which are beyond those typically used during hovering, suggesting the model remains valid within the robot’s intended operational flap angles.
%
\subsection*{Force analysis}
Our robot consists of multiple rigid bodies. Therefore, the total force measured on the robot is the sum of the gravitational force and the individual drag forces ${F}_\mathrm{D}$ exerted on each component of the robot's body (the base ${F}_\mathrm{B}$ and the four flaps ${F}_{i}$) as shown in equation (\ref{eq:total_force}).\\
The gravitational force acts on the center of mass, which is designed to be in the center of the base with a tunable vertical shift. The flaps have low weight and are designed with their center of mass located close to their rotation axes, ensuring that changing their configuration does not affect the position of the robot's center of mass.
Given that the base is designed as a thin plate, we use the simplified drag force model and write the drag force acting on the base as equation (\ref{eq:base_drag_force}):
\begin{equation}
    {F}_{\mathrm{B}} = \frac{1}{2} \rho C_{\mathrm{D,B}} A_\mathrm{B} \left(({v}_\mathrm{air} - {v}_\mathrm{B})^{\top} {e}_\mathrm{z,B} \right)^2 {e}_\mathrm{z,B},
    \label{eq:base_drag_force}
\end{equation}
where $\rho$ is the air density, $C_{\mathrm{D,B}}$ is the drag coefficient of the base, $A_{\mathrm{B}}$ is the area of the base, $v_{\mathrm{air}}$ is the airflow speed, $v_{\mathrm{B}}$ is the velocity of the base, and ${e}_\mathrm{z,B}$ is the $\mathrm{z}$ vector of the body's coordinate frame. Similarly, the flaps are designed as a combination of flat plates. Hence, the drag force on flap $F_i$ can be written as equation (\ref{eq:flaps_drag_force}):
\begin{equation}
    {F}_{i} = \frac{1}{2} \rho \sum_{j=1}^n C_{\mathrm{D,F}_j} A_{\mathrm{F}_j} \left(({v}_\mathrm{air} - {v}_{i,j})^{\top} {n}_{j} \right)^2  {n}_{j},
    \label{eq:flaps_drag_force}
\end{equation}
where $n$ is the number of different flat plates in each flap (in our case two), $C_{\mathrm{D,F}_j}$ is the drag coefficient of the $j^{th}$ plate of the flap, $A_{\mathrm{F}_j}$ is its area, ${v}_{i,j}$ is its velocity vector, and ${n}_{j}$ is its normal direction.
In the case of single plate flaps, the equation is reduced as shown in equation (\ref{eq:flaps_drag_force_simplified}):
\begin{equation}
    {F}_{i} = \frac{1}{2} \rho C_{\mathrm{D,F}} A_{\mathrm{F}} \left(({v}_\mathrm{air} - {v}_{i})^{\top} {e}_{\mathrm{z},i} \right)^2  {e}_{\mathrm{z},i},
    \label{eq:flaps_drag_force_simplified}
\end{equation}
where $v_i$ is the flap's velocity vector, and ${e}_\mathrm{z,i}$ is the $\mathrm{z}$ vector of the $i\mathrm{th}$ flap coordinate frame. This model calculates the drag force as if each plate is isolated in the airflow, unaffected by the presence of other plates. Despite this simplification, the theoretical results closely align with the experimental data.
\subsection*{Torque analysis}
The torque acting on the center of mass of the robot has two contributing factors $T_\mathrm{T} = T_\mathrm{D} + T_\mathrm{s}$. The primary factor is the torque generated by the drag forces $T_\mathrm{D}$, which can be calculated as shown in equation (\ref{eq:drag_torque_equation}):
\begin{equation}
    T_\mathrm{D} = \sum^{m}_{i=1} r_i\times F_{i},
    \label{eq:drag_torque_equation}
\end{equation}
where $m$ is the number of drag forces $F_{i}$ is the $i^\mathrm{th}$ drag force, and $r_i$ is the distance between the center of mass and the center of the pressure where the drag force $F_{i}$ is applied, which represents the lever arm. The second factor $T_\mathrm{s}$ is the counter-torque produced by the servo motors when moving the flaps as described in equation (\ref{eq:motors_torque}):
\begin{equation}
    T_\mathrm{s} = I_\mathrm{f} \ddot{\alpha}_\mathrm{s},
    \label{eq:motors_torque}
\end{equation}
where $I_\mathrm{f}$ is the moment of inertia of the flap, and $\ddot{\alpha}_\mathrm{s}$ is the servo's angular acceleration. However, since the flaps are designed to be lightweight and have therefore small inertia around their axis of rotation, this counter-torque will be omitted. As a result, the total torque is simplified to the drag-induced torque alone $T_\mathrm{T} \approx T_\mathrm{D}$. These forces and torques, directly affect the robot's dynamics as detailed in the following sections.

\subsection*{High level analysis of the actuation principle}
Even though the control inputs directly affect the flap angles, rotating a single flap results in a compound effect on the force and torque measured on the robot.
This occurs because rotating a flap affects both the magnitude of the drag force acting on it and the direction in which the force is applied.
Thus, we take advantage of the robot's symmetry to design orthogonal compound control signals  $\Tilde{u}_i$, each affecting only one of the targeted control quantities ($x$, $y$, $z$-torques, and $z$-force).\\
The actuation principle can be explained without the need to analyze the drag force on each flap. We see that the robot is symmetrical about the $xz$-plane, and about the $yz$-plane. Hence, assuming homogenous airflow, if the drag force acting on flap one $F_{\mathrm{f}1}$ is written: $F_{\mathrm{f}1} = [F_\mathrm{x}, F_\mathrm{y}, F_\mathrm{z}]^{\top}$, then the other forces can be written as follows $F_{\mathrm{f}2} = [F_\mathrm{x}, -F_\mathrm{y}, F_\mathrm{z}]^{\top}$, $F_{\mathrm{f}3} = [-F_\mathrm{x}, -F_\mathrm{y}, F_\mathrm{z}]^{\top}$, and $F_{\mathrm{f}4} = [-F_\mathrm{x}, F_\mathrm{y}, F_\mathrm{z}]^{\top}$. Assuming that in the hover state, the robot's base is horizontal, and its drag force has a $z$ component only, we can conclude the following:\\
1- In the hover configuration, the total drag force has only a $z$ component and can be tuned to counter gravity as shown in equation (\ref{eq:total_force_z_component}):
\begin{equation}
    {F}_\mathrm{B} + \Sigma_{i=1}^{4}{F}_{\mathrm{f}i} = [0, 0, \mathrm{mg}]^{\top},
    \label{eq:total_force_z_component}
\end{equation}
where $\mathrm{m}$ denotes the robot's mass and $\mathrm{g}$ the gravitational acceleration.\\
2- Due to the symmetrical configuration, all torques cancel out.
Assuming that at hover configuration $\hat{\theta} = (\theta_\mathrm{H}, -\theta_\mathrm{H}, \theta_\mathrm{H}, -\theta_\mathrm{H})$, the drag force acting on flap one, is described by equation (\ref{eq:drag_force_flap_one_hover}):
\begin{equation}
    F_{\mathrm{f}1}(\theta_\mathrm{H}) = [F_\mathrm{x,H}, F_\mathrm{y,H}, F_\mathrm{z,H}]^{\top},
    \label{eq:drag_force_flap_one_hover}
\end{equation}
then for a small deviation in the flap angle $\Delta\theta$, the drag force can be linearized as given in equation (\ref{eq:drag_force_flap_one_hover_deviation}):
\begin{equation}
    F_{\mathrm{f}1}(\theta_\mathrm{H}+\Delta\theta) \approx F_{\mathrm{f}1}(\theta_\mathrm{H})+\Delta\theta \frac{\partial F_{\mathrm{f}1}}{\partial\theta}(\theta_\mathrm{H}) = 
    \begin{bmatrix}
        F_\mathrm{x,H} +\Delta \theta F'_\mathrm{x,H} \\
        F_\mathrm{y,H} +\Delta \theta F'_\mathrm{y,H} \\
        F_\mathrm{z,H} +\Delta \theta F'_\mathrm{z,H}
    \end{bmatrix}.
    \label{eq:drag_force_flap_one_hover_deviation}
\end{equation}
From symmetry and considering that at the hover configuration, the flaps 1 and 3 have an angle \{$\theta_\mathrm{H}$\}, while flaps 2 and 4 have an angle \{$-\theta_\mathrm{H}$\}, we can conclude that values for $\frac{\partial F_{\mathrm{f}2}}{\partial\theta}(-\theta_\mathrm{H})$, $\frac{\partial F_{\mathrm{f}3}}{\partial\theta}(\theta_\mathrm{H})$, $\frac{\partial F_{\mathrm{f}4}}{\partial\theta}(-\theta_\mathrm{H})$ can be expressed as detailed in equation (\ref{eq:drag_force_flaps_derivative}):
\begin{equation}
    \frac{\partial F_{\mathrm{f}1}}{\partial\theta}(\theta_\mathrm{H}) = 
    \begin{bmatrix}
        F'_\mathrm{x,H} \\
        F'_\mathrm{y,H} \\
        F'_\mathrm{z,H}
    \end{bmatrix},~
    \frac{\partial F_{\mathrm{f}2}}{\partial\theta}(-\theta_\mathrm{H}) = 
    \begin{bmatrix}
        -F'_\mathrm{x,H} \\
        F'_\mathrm{y,H} \\
        -F'_\mathrm{z,H}
    \end{bmatrix},~
    \frac{\partial F_{\mathrm{f}3}}{\partial\theta}(\theta_\mathrm{H}) = 
    \begin{bmatrix}
        -F'_\mathrm{x,H} \\
        -F'_\mathrm{y,H} \\
        F'_\mathrm{z,H}
    \end{bmatrix},~
    \frac{\partial F_{\mathrm{f}4}}{\partial\theta}(-\theta_\mathrm{H}) = 
    \begin{bmatrix}
        F'_\mathrm{x,H} \\
        -F'_\mathrm{y,H} \\
        -F'_\mathrm{z,H}
    \end{bmatrix}.
    \label{eq:drag_force_flaps_derivative}
\end{equation}
We now leverage these symmetry conditions to analyze the roll, pitch, yaw, and height dynamics in the following paragraphs.
%
\subsubsection*{Z force control}
The vertical force can be controlled with $\tilde{u}_4$ by closing or opening the four flaps simultaneously with the same angle. If we consider the compound control signal $\tilde{u} = [0, 0, 0, -\Delta\theta]$, then according to equation (\ref{eq:input_signals}), the flaps' angles deviate from the hover configuration as following $\hat{\theta} = (\theta_\mathrm{H} + \Delta\theta, -\theta_\mathrm{H} - \Delta\theta, \theta_\mathrm{H} + \Delta\theta, -\theta_\mathrm{H} - \Delta\theta)$
and similar to equation (\ref{eq:total_force}), lateral forces cancels out and the total force has only a $z$ component as shown in equation (\ref{eq:total_force_z_control}):
\begin{equation}
    {F}_\mathrm{B} + \Sigma_{i=1}^{4}{F}_{\mathrm{f}i}(\hat{\theta}_i) = [0, 0, \mathrm{mg} + 4 \Delta \theta F'_\mathrm{z,H}]^{\top}.
    \label{eq:total_force_z_control}
\end{equation}
Because symmetry in the flap configuration is maintained, all torques cancel out, which allows for independent control of the vertical dynamics.
\subsubsection*{Yaw torque control}
The yaw torque can be controlled with $\tilde{u}_3$ by simultaneously closing two opposite flaps and opening the other two by the same angle, as shown in Fig.~\ref{fig:control_concept}~b. If we consider the compound control signal $\tilde{u} = [0, 0, -\Delta\theta, 0]$, then according to equation (\ref{eq:input_signals}), the flaps' angles deviate from the hover configuration as follows $\hat{\theta} = (\theta_\mathrm{H} + \Delta\theta, -\theta_\mathrm{H} + \Delta\theta, \theta_\mathrm{H} + \Delta\theta, -\theta_\mathrm{H} + \Delta\theta)$.
In this configuration, the lateral forces cancel out, and the total force remains unchanged with only a $z$ component equal to the robot's weight, as shown in equation (\ref{eq:total_force}). However, the nonsymmetrical flap angles result in a net yaw torque, as the $z$-axis torques no longer cancel out as described in equation (\ref{eq:total_torque_yaw_control}):
\begin{equation}
    \Sigma_{i=1}^{4}{T}_{\mathrm{f}i}(\hat{\theta}_i) = [0, 0, -4 \Delta \theta |\mathrm{dC_{p,xy}}\times F'_\mathrm{xy,H}|^2]^{\top}.\,
    \label{eq:total_torque_yaw_control}
\end{equation}
where $\mathrm{dC_{p,xy}}$ is the horizontal $xy$ component of the lever arm vector, which extends from the center of mass to the center of pressure on flap “1” (where the drag force is applied), $F'_\mathrm{xy,H}$ is the horizontal $xy$ component of the change in the drag force vector experienced on flap “1”, i.e. $ F'_\mathrm{xy,H} = [F'_\mathrm{x,H}, F'_\mathrm{y,H}, 0]^{\top}$, and $| \cdot |$ denotes the Euclidean norm.
\subsubsection*{Roll and pitch control}
Similar to controlling yaw, controlling roll and pitch can be achieved by inducing asymmetry in the flaps configuration with $\tilde{u}_1$ and $\tilde{u}_2$, which close two adjacent flaps and open the other two as can be seen in Fig.~\ref{fig:control_concept}~c. If we consider the compound control signal $\tilde{u} = [\Delta\theta, 0, 0, 0]$, then according the equation (\ref{eq:input_signals}), the flaps' angles deviate from the hover configuration as follows $\hat{\theta} = (\theta_\mathrm{H} + \Delta\theta, -\theta_\mathrm{H} + \Delta\theta, \theta_\mathrm{H} - \Delta\theta, -\theta_\mathrm{H} - \Delta\theta)$
. Similar to the yaw case, lateral forces cancels out and the total force has only a $z$ component equal to the mass of the robot. However, due to the nonsymmetrical flap angles, the $x$ “roll” torque does not cancel out as shown in Fig.~\ref{fig:control_concept}~c and described in equation (\ref{eq:total_torque_roll_control}):
\begin{equation}
    \Sigma_{i=1}^{4}{T}_{\mathrm{f}i}(\hat{\theta}_i) = [4 \Delta \theta |\mathrm{dC_{p,yz}}\times F'_\mathrm{yz,H}|^2, 0, 0]^{\top},
    \label{eq:total_torque_roll_control}
\end{equation}
where $\mathrm{dC_{p,yz}}$ is the $yz$ component of the lever vector, and $F'_\mathrm{yz,H}$ is the $yz$ component of the change in the drag force vector experienced on flap “1” $ F'_\mathrm{yz,H} = [0, F'_\mathrm{y,H}, F'_\mathrm{z,H}]^{\top}$.\\
In a similar manner, the pitch torque caused by the input signal $\tilde{u} = [0, \Delta\theta, 0, 0]$ is described in equation (\ref{eq:total_torque_pitch_control}):
\begin{equation}
    \Sigma_{i=1}^{4}{T}_{\mathrm{f}i}(\hat{\theta}_i) = [0, 4 \Delta \theta |\mathrm{dC_{p,xz}}\times F'_\mathrm{xz,H}|^2, 0]^{\top},
    \label{eq:total_torque_pitch_control}
\end{equation}
where $\mathrm{dC_{p,xz}}$ is the $xz$ component of the lever vector, and $F'_\mathrm{xz,H}$ is the $xz$ component of the change in the drag force vector experienced on flap 1, i.e. $ F'_\mathrm{xz,H} = [F'_\mathrm{x,H}, 0,  F'_\mathrm{z,H}]^{\top}$.\\
These equations collectively describe the influence of input signals on the height, roll, pitch, and yaw dynamics of the system. When combined with the simplified drag model, the equations enable the computation and prediction of the structural and input constants of the design.
\subsection*{Setup specifications}
\begin{figure} 
	\centering
	\includegraphics[width=1.0\textwidth]{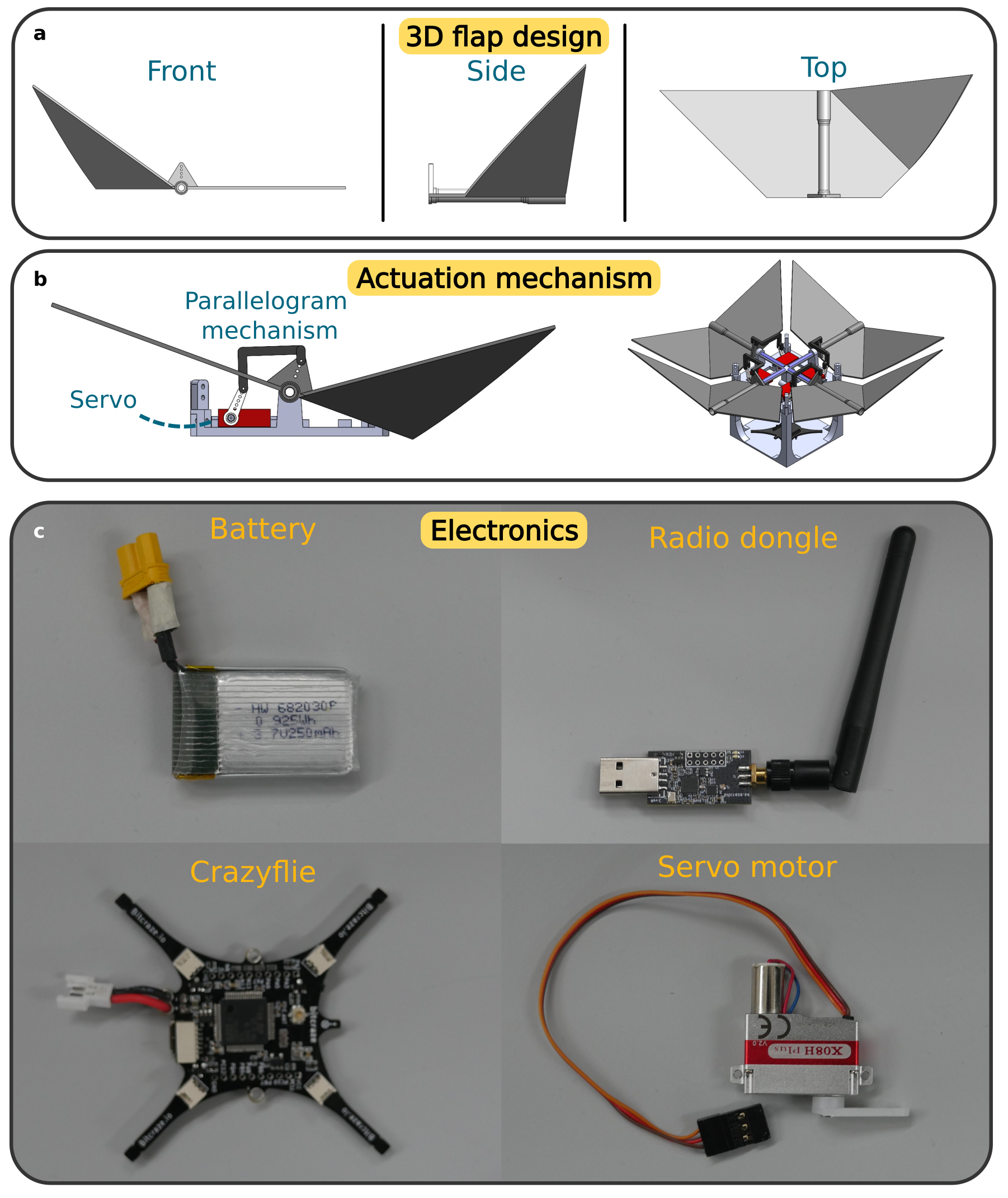} 
	\caption{\textbf{Design and used electronics.}
		(\textbf{a}) The flaps have a kink on one side angled at $42.5^\circ$. (\textbf{b}) The servo motors rotate the flaps via a parallelogram mechanism. (\textbf{c}) The robot uses the Crazyflie controller, a single-cell LiPo $250~\mathrm{mAh}$ battery, and four KST servos. A radio dongle is used for the communication with the robot.}
	\label{fig:design_and_electronics} 
\end{figure}

\textbf{Robot:}
The robot consists of the Crazyflie flight controller, two $9~\mathrm{g}$ batteries, a voltage regulation circuit, and four KST-X08 Plus servo motors which are connected to the flaps with a parallelogram mechanism Fig.~\ref{fig:design_and_electronics}. Eight reflective markers are attached to the robot for tracking its position and orientation using an OptiTrack motion capture system. Both the robot's body and the flaps are 3D printed. The body frame is printed using fiber-reinforced filament (Onyx, Markforged) with a base thickness of $3~\mathrm{mm}$. The flaps are printed with PLA Tough filament (Bambu Lab), each having a thickness of  $1.6~\mathrm{mm}$ and weighing approximately $25~\mathrm{g}$. Collectively, the four flaps contribute to about one-third of the robot's total weight ($100~\mathrm{g}$). The flaps have trapezoidal shape with kinks on one side angled at $42.5^\circ$ Fig.~\ref{fig:design_and_electronics}~a. Lightweight carbon-fiber rods, $4~\mathrm{mm}$ in diameter, are used as rotation axes for the flaps. The robot’s maximum footprint is $23~\mathrm{cm}\times23~\mathrm{cm} $ with a total mass of $340~\mathrm{g}$.
\\
\textbf{Motion capture system:}
For tracking the robot's position and orientation, we use five OptiTrack cameras, two “$\mathrm{Prime^x}$ 13” and three “$\mathrm{Prime^x}$ 13W”. The OptiTrack tracks the robot with $200~ \mathrm{Hz}$ using eight reflective markers positioned on the robot. The OptiTrack system then send the information to the off-board system which calculates the robot's error in position and orientation and sends the correction information to the robot via the Crazyflie radio dongle with a frequency of $50~\mathrm{Hz}$. With the same frequency, the off-board system receives and logs information and sensory data from the robot. The OptiTrack cameras cover the area over the wind tunnel up to $2~\mathrm{m}$ above the wind tunnel.
\\
\textbf{Wind tunnel:}
We use a custom-built wind tunnel that is designed and built specifically to facilitate the experiments for this project. The wind tunnel has $1.2~\mathrm{m}$ diameter and uses seven U5 KV400 brushless motors equipped with $406~\mathrm{mm}$ diameter T-motor carbon fiber propellers to generate a vertical airflow with average velocities ranging from $8~\mathrm{m/s}$ to $11~\mathrm{m/s}$. Six motors are positioned at the corners of a regular hexagon with a side length of $28~\mathrm{cm}$ with the seventh motor positioned at the center of the wind tunnel. The frame of the wind tunnel is constructed with $40\times40~\mathrm{mm}$ aluminum profiles. Three plexiglass plates are used to create a cylindrical shield around the wind tunnel and acts as a funnel for the air. A $14~\mathrm{cm}$ honeycomb layer is used above the propellers to straighten the airflow. 
\subsection*{Experimental setup}
Due to the fact that our wind tunnel is a custom-made wind tunnel, the airflow is neither laminar nor homogeneous. Therefore, all experiments are performed in a cylindrical area with a diameter of about $30~\mathrm{cm}$ and height of about $40~\mathrm{cm}$ located at $100~\mathrm{cm}$ above the wind tunnel surface. For safety measures, the robot is connected with a fishing thread to ensure that the robot cannot fall outside the wind tunnel and get damaged. The thread is connected to a small counter-weight via a double pulley system ensuring the thread retracts slightly during flight and does not interfere.
\subsection*{Detailed derivation of scaling analysis}
\textbf{Size scaling:}
We investigate the influence of the robot's scale on performance and energy efficiency. Our model does not explicitly account for energy consumption, and therefore two proxy metrics are analyzed: (1) the normalized linear acceleration-to-input ratio, ${\ddot{\tilde{x}}}/{\tilde{u}}$ where $\ddot{\tilde{x}} =\ddot{x}/{l}$ (linear acceleration normalized by the robot's side length $l$) and (2) the angular acceleration-to-input ratio, ${\dot{\omega}}/{\tilde{u}}$. These metrics provide insight into the robot's agility, describing the input, and therefore the energy required to achieve given relative accelerations.
The analysis assumes that the robot's hover angle remains constant across scales, necessitating adjustments to airflow velocity. 
The results indicate that reducing the robot’s scale enhances agility. Specifically, while the absolute linear acceleration-to-input ratio is invariant under scaling, the normalized linear acceleration-to-input ratio increases. Similarly, the angular acceleration-to-input ratio increases with downscaling. Both metrics exhibit an inverse proportionality to the side length of the robot, expressed as ${\ddot{\tilde{x}}}/{\tilde{u}} \propto {1}/{l}$ and ${\dot{\omega}}/{\tilde{u}} \propto {1}/{l}$. This implies that halving the robot’s side length effectively doubles its agility.\\
This behavior can be attributed to the interplay between drag and gravitational forces during hovering. Drag counteracts gravity to produce an upward acceleration equal to $g$. Applying a specific input $\tilde{u}$ modifies the drag forces on the flaps, altering their magnitude and direction to produce a linear acceleration, $\ddot{x}=f(g,\tilde{u})$, which is independent of scale under the given assumptions. Consequently, normalized linear acceleration ${\ddot{x}}/{l}$, scales inversely with $l$. Angular acceleration behaves similarly; although torque scales proportionally to $l$ due to the lever arm’s dependence on $l$, the moment of inertia scales with $l^2$. As a result, the angular acceleration scales inversely with $l$. This suggests that scaling down the robot enhances its agility, and thus its energy efficiency, as smaller inputs suffice for operation.\\
\textbf{Mass scaling:}
It is important to note that the aforementioned scale-agility relationship is independent of the robot’s mass under the assumption of a constant hover angle. Although mass scaling influences forces and torques, it does not affect accelerations. This allows the robot's mass to be tuned to adapt to specific airflow conditions in different environments.
While mass scaling does not directly influence agility, it affects the airflow speed required to achieve hover. The drag force, $F_\mathrm{d}$, balances the gravitational force, $mg$, and is expressed in equation (\ref{eq:drag_force_proportionality}):
\begin{equation}
    F_\mathrm{d} = mg \propto l^2 v^2_\mathrm{air}.
    \label{eq:drag_force_proportionality}
\end{equation}
From this relationship, the required airflow speed is proportional to the square root of the robot’s mass divided by its side length squared as shown in equation (\ref{eq:airflow_proportionality})
Different airflow speeds result in varying levels of disturbances, which can affect the robot’s performance and energy consumption. To better understand this, we analyze the effect of mass scaling on the Reynolds number for the airflow experienced by the robot (in our case $\mathrm{R_e}\approx2\times10^5$). The Reynolds number, while not providing a complete description of airflow behavior, serves as a useful indicator of laminar or turbulent flow conditions \cite{shyy2008aerodynamics, wang2004unsteady}. The Reynolds number is influenced by the characteristic length (the robot’s side length $l$) and the airflow speed $v_\mathrm{air}$, and is expressed in equation (\ref{eq:Reynold_numbers});
where $\rho$ is the air density, and $\mu$ is the dynamic viscosity of air. This indicates that while scaling the robot's size does not affect the Reynolds number, scaling the mass directly impacts it. Specifically, reducing the robot’s mass results in a lower Reynolds number, potentially favoring more laminar flow.
In conclusion, reducing Floaty's size enhances agility and energy efficiency by increasing normalized acceleration-to-input ratios, allowing more responsive maneuvers with lower power consumption. Additionally, decreasing mass lowers the required airflow speed and consequently, the Reynolds number, promoting efficient flight. These findings suggest that a smaller, lighter design optimizes flight performance, making Floaty ideal for dynamic vertical airflow environments.


\subsection*{Estimation and calculation of specific power consumption for aerial vehicles}
\label{sec:si_spc_calculation}

This section details the methodology employed to estimate and calculate the specific power consumption (SPC) for the diverse range of aerial vehicles included in our comparative analysis (Fig.~\ref{fig:robustness}~a~and~b in main text). SPC, defined as the power required for hover per unit of takeoff mass (expressed in Watts per kilogram, $\mathrm{W/kg}$), serves as a key metric for evaluating the energetic efficiency of vertical flight.
\subsubsection*{Multirotor drones}
\label{ssec:spc_drones}

For commercially available multirotor drones (e.g., DJI Air 3, DJI Mini 4 Pro, DJI Avata, Bitcraze Crazyflie 2.1) and the custom-built quadcopter, hovering power ($P_{\mathrm{hover}}$, in Watts) is estimated from manufacturer-supplied battery specifications and maximum flight endurance data. The total energy capacity of the onboard battery ($E_{\mathrm{batt}}$, in Watt-hours) is obtained either directly from the manufacturer's stated Watt-hour ($\mathrm{Wh}$) rating or calculated by multiplying the nominal battery voltage ($V_{\mathrm{nom}}$, in Volts) by its charge capacity ($C_{\mathrm{batt}}$, in Ampere-hours, $\mathrm{Ah}$). Manufacturer-specified maximum flight times ($T_{\mathrm{flight}}$, in hours) -- typically quoted under ideal, no-wind, sea-level conditions and often achieved during hover or very gentle flight \cite{dji_air3_specs_2023, dji_avata_specs_2022, dji_mavic3pro_specs_2023, dji_mini4pro_specs_2023, giernacki2017crazyflie} -- are used as a proxy for sustained hover endurance. The average power consumption, assumed to represent $P_{\mathrm{hover}}$, is then derived using Equation~\ref{eq:si_phover}:
\begin{equation}
P_{\mathrm{hover}} = \frac{E_{\mathrm{batt}}}{T_{\mathrm{flight}}}
\label{eq:si_phover}
\end{equation}
The takeoff mass ($M_{\mathrm{takeoff}}$, in kilograms) for these drones is sourced directly from official manufacturer specifications \cite{dji_air3_specs_2023, dji_avata_specs_2022, dji_mini4pro_specs_2023, giernacki2017crazyflie}.

\subsubsection*{Conventional helicopters}
\label{ssec:spc_helicopters}

Direct electrical-equivalent hovering power consumption figures are not published for conventional helicopters (e.g., Robinson R22 Beta II, Bell 206B JetRanger, Airbus H125, Sikorsky UH-60M Black Hawk). Therefore, $P_{\mathrm{hover}}$ is estimated based on the engine(s)' shaft horsepower (shp) required for out of ground effect hover. This required power is approximated as a percentage of the engine(s)' maximum continuous power rating, a common practice in preliminary performance estimation \cite{prouty_helicopter_2005}. The specific percentage varied based on helicopter class and engine type: typically 75--90\% for piston-engine helicopters (e.g., Robinson R22) due to their operation closer to power limits during hover, and 60--75\% for turbine-engine helicopters (e.g., Bell 206B, Airbus H125, UH-60M), reflecting their generally higher power reserves \cite{prouty_helicopter_2005, leishman2006principles}. Engine power ratings (in shp) and maximum takeoff weights ($M_{\mathrm{takeoff}}$) are obtained from official manufacturer documentation, publicly available type certificate data sheets from aviation authorities such as the U.S. Federal Aviation Administration \cite{faa_tcds_h2sw_bell206}. Shaft horsepower is converted to Watts using the standard conversion factor (1 shp $\approx$ 745.7 $\mathrm{W}$).

\subsubsection*{Calculation of specific power consumption}
\label{ssec:spc_calculation_final}

For all vehicles, once $P_{\mathrm{hover}}$ ($\mathrm{W}$) and $M_{\mathrm{takeoff}}$ ($\mathrm{kg}$) are determined or estimated as described above, the hover specific power consumption ($P_{\mathrm{specific}}$) is calculated using Equation~\ref{eq:si_pspecific}:
\begin{equation}
P_{\mathrm{specific}} \mathrm{ (W/kg)} = \frac{P_{\mathrm{hover}} \mathrm{ (W)}}{M_{\mathrm{takeoff}} \mathrm{ (kg)}}.
\label{eq:si_pspecific}
\end{equation}

\subsubsection*{Data for updraft conditions}
\label{ssec:spc_updraft}

For the Crazyflie 2.1 and the custom quadcopter, power consumption under updraft conditions ($P_{\mathrm{updraft}}$) is measured empirically in a controlled wind tunnel environment generating an approximate 10 m/s vertical updraft. The specific power consumption in updraft is then calculated as $P_{\mathrm{updraft}} \mathrm{ (W)} / M_{\mathrm{takeoff}} \mathrm{ (kg)}$.

\subsubsection*{Limitations and assumptions}
\label{ssec:spc_limitations}

It is important to acknowledge that the specific power consumption values presented, particularly for conventional helicopters and electrical vehicles, are estimations based on the best available public data and established engineering approximations. Actual power consumption during flight is a complex function of numerous variables, including atmospheric conditions (density altitude), payload, specific flight maneuvers, individual aircraft condition, and pilot technique. The methodology adopted aims to provide a standardized basis for comparison across multiple vehicle classes and scales, assuming ideal or representative hover conditions as defined by the data sources.


\newpage



\begin{figure} 
	\centering
	\includegraphics[width=1.0\textwidth]{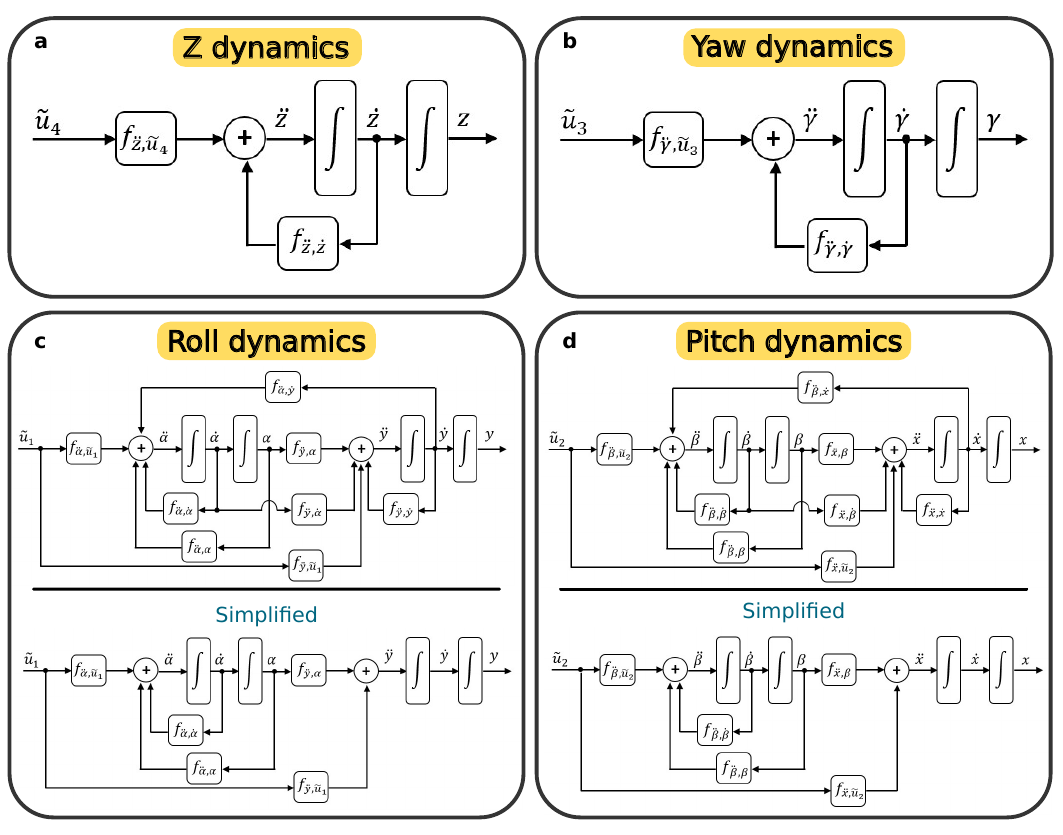} 
	\caption{\textbf{Decoupled dynamics of the robot.}
		(\textbf{a}) The input $\Tilde{u}_4$ controls the height ($z$) dynamics. Similarly, (\textbf{b}) the input $\Tilde{u}_3$ controls the yaw dynamics. The inputs $\Tilde{u}_1$ (\textbf{c}) and $\Tilde{u}_2$ (\textbf{d}) control the roll and pitch dynamics, respectively. Both roll and pitch dynamics can be simplified by neglecting small values of the structural constants. The values of the input and structural constants can be found in Table \ref{tab:dynamic_consts}.}
	\label{fig:decoupled_dynamics} 
\end{figure}

\begin{table}
	\centering

    	\caption{\textbf{Empirical estimation of dynamics constants.}
		The structural and input constants for the robot dynamics.}
	\label{tab:dynamic_consts}

	\begin{tabular}{cccc} 
            \\
		\hline
		constant & value & units & comment \\
		\hline
		$f_{\ddot{\alpha}, \Tilde{u}_1}$ & $230 \pm 66$ & $\mathrm{s}^{-2}$ & estimated using flight data \\
		$f_{\ddot{\alpha}, \alpha}$ & $-16 \pm 10$ & $\mathrm{s}^{-2}$ & estimated using flight data \\
		$f_{\ddot{\alpha}, \dot{\alpha}}$ & $-40 \pm 12$ & $\mathrm{s}^{-1}$ & estimated using flight data \\
		$f_{\ddot{y}, \Tilde{u}_1}$ & $-4.8 \pm 0.1$ & $\mathrm{m~s}^{-2}\mathrm{rad}^{-1}$ & estimated using force sensor measurements \\
		$f_{\ddot{y}, \alpha}$ & $-6.5 \pm 0.1$ & $\mathrm{m~s}^{-2}\mathrm{rad}^{-1}$ & estimated using physical model \\
		$f_{\ddot{\beta}, \Tilde{u}_2}$ & $1100 \pm 170$ & $\mathrm{s}^{-2}$ & estimated using flight data \\
		$f_{\ddot{\beta}, \beta}$ & $-1560 \pm 25$  & $\mathrm{s}^{-2}$ & estimated using flight data \\
		$f_{\ddot{\beta}, \dot{\beta}}$ & $-29 \pm 5$ & $\mathrm{s}^{-1}$ & estimated using flight data \\
		$f_{\ddot{x}, \Tilde{u}_2}$ & $-0.8 \pm 0.1$ & $\mathrm{m~s}^{-2}\mathrm{rad}^{-1}$ & estimated using force sensor measurements \\
		$f_{\ddot{x}, \beta}$ & $2.5 \pm 0.1$ & $\mathrm{m~s}^{-2}\mathrm{rad}^{-1}$ & estimated using physical model \\
		$f_{\ddot{\gamma}, \Tilde{u}_3}$ & $110 \pm 1.8$ & $\mathrm{s}^{-2}$ & estimated using flight data \\
		$f_{\ddot{\gamma}, \dot{\gamma}}$ & $-7.7 \pm 0.1$ & $\mathrm{s}^{-1}$ & estimated using flight data \\
		$f_{\ddot{z}, \Tilde{u}_4}$ & $2.3 \pm 0.3$ & $\mathrm{m~s}^{-2}\mathrm{rad}^{-1}$ & estimated using force sensor measurements \\
		$f_{\ddot{z}, \dot{z}}$ & $-2 \pm 0.1$ & $\mathrm{s}^{-1}$ & estimated using physical model \\
		\hline
	\end{tabular}
\end{table}



\clearpage 

\paragraph{Supplementary videos 1-7 can be accessed via the link: \href{https://ghadeerelmkaiel.github.io/Floaty-Project/}{https://ghadeerelmkaiel.github.io/Floaty-Project/}}

\paragraph{Supplementary \href{https://ghadeerelmkaiel.github.io/Floaty-Project/}{Video} 1:}
\textbf{Floaty.}
The robot's design and some experiment in our setup Fig.~\ref{fig:main_figure}.

\paragraph{Supplementary \href{https://ghadeerelmkaiel.github.io/Floaty-Project/}{Video} 2:}
\textbf{Control Concept.}
A visualization of our actuation principle to illustrate the effect of the input on the dynamics. More details can be seen in Fig.~\ref{fig:control_concept}.

\paragraph{Supplementary \href{https://ghadeerelmkaiel.github.io/Floaty-Project/}{Video} 3:}
\textbf{Hover experiment.}
The robot performs a static target tracking, shown in red, with the real-time error in position and orientation displayed at the bottom. More data can be seen in Fig.~\ref{fig:Hover_figure}.

\paragraph{Supplementary \href{https://ghadeerelmkaiel.github.io/Floaty-Project/}{Video} 4:}
\textbf{Height tracking experiment.}
The robot performs a dynamic target tracking with the target position moving vertically, following a sine wave, shown in red. The error in position and the $z$-control command are displayed to the right. More data can be seen in Fig.~\ref{fig:experiments_fig}~c.

\paragraph{Supplementary \href{https://ghadeerelmkaiel.github.io/Floaty-Project/}{Video} 5:}
\textbf{Yaw tracking experiment.}
The robot performs a yaw angle tracking with the target orientation following a sine wave with an amplitude of $\pi/2$. The target orientation is illustrated with a red arrow. More data can be seen in Fig.~\ref{fig:experiments_fig}~a,~b.

\paragraph{Supplementary \href{https://ghadeerelmkaiel.github.io/Floaty-Project/}{Video} 6:}
\textbf{Quadcopter flight in a wind tunnel.}
Comparison of the flight performance of two different scales of quadcopters with and without vertical airflow. Micro Crazyflie $27~\mathrm{g}$ quadcopter, and a custom-built $40\times40~\mathrm{cm}$, $940~\mathrm{g}$ quadcopter.

\paragraph{Supplementary \href{https://ghadeerelmkaiel.github.io/Floaty-Project/}{Video} 7:}
\textbf{Robustness and disturbance rejection.}
The robot is capable of maintaining flight across multiple experiments with different disturbances.



\end{document}